\documentclass{article}

\usepackage{PRIMEarxiv}

\usepackage[utf8]{inputenc} 
\usepackage[T1]{fontenc}    
\usepackage{hyperref}       
\usepackage{url}            
\usepackage{booktabs}       
\usepackage{amsfonts}       
\usepackage{nicefrac}       
\usepackage{microtype}      
\usepackage{lipsum}
\usepackage{fancyhdr}       
\usepackage{graphicx}       
\usepackage{amsmath,amssymb} 
\usepackage{mathtools}
\usepackage{graphicx}
\usepackage{subfigure}
\usepackage{tikz}
\usepackage{comment}
\usepackage{color}
\usepackage{booktabs}
\usepackage[ruled,linesnumbered]{algorithm2e}
\usepackage[title]{appendix}
\usepackage{multirow}

\graphicspath{{media/}}     

\DeclareMathOperator*{\argmin}{argmin}
\DeclareMathOperator*{\argmax}{argmax}
\title{A Simple Test-Time Method for Out-of-Distribution Detection
}

\author{
  Ke Fan \\
  School of Data Science \\
  Fudan University \\
  Shanghai\\
  \texttt{kfan21@m.fudan.edu.cn} \\
   \And
  Yikai Wang \\
  School of Data Science \\
  Fudan University \\
  Shanghai\\
  \texttt{yikaiwang19@fudan.edu.cn} \\
   \And
  Qian Yu \\
  Beihang University \\
  Beijing\\
  \texttt{qianyu@buaa.edu.cn} \\
   \And
  Da Li \\
  Samsung AI Center \\
  Cambridge\\
  \texttt{dali.academic@gmail.com} \\
   \And
  Yanwei Fu \\
  School of Data Science \\
  Fudan University \\
  Shanghai\\
  \texttt{yanweifu@fudan.edu.cn} \\
}

\begin{document}
\maketitle

\begin{abstract}
Neural networks are known to produce over-confident predictions on input images, even when these images are out-of-distribution (OOD) samples.
This limits the applications of neural network models in real-world scenarios, where OOD samples exist.
Many   existing  approaches identify the OOD instances via exploiting various cues, such as  finding irregular patterns in the feature space, logits space, gradient space or the raw space of images. In contrast, this paper proposes a simple Test-time Linear Training (ETLT)   method for OOD detection. Empirically,  we find that the probabilities of input images being out-of-distribution are surprisingly  linearly correlated to the features extracted by neural networks.
To be specific, many state-of-the-art OOD algorithms, although designed to measure reliability in different ways, actually lead to OOD scores 
mostly linearly related to their image features. Thus, by simply learning a  linear regression model trained from the paired image features and inferred OOD scores at test-time,  we can make a more precise OOD prediction for the test instances.
We further propose an online variant of the proposed method, which achieves promising performance and is more practical in real-world applications.
Remarkably, we improve FPR95 from $51.37\%$ to $12.30\%$ on CIFAR-10 datasets with maximum softmax probability as the base OOD detector.
Extensive experiments on several benchmark datasets show the efficacy of ETLT  for OOD detection task.
\end{abstract}

\section{Introduction}
\label{sec:intro}
Deep neural networks are known to achieve outstanding performance on image recognition tasks~\cite{simonyan2014very,he2015deep,huang2017densely,dosovitskiy2020image}.
However, the model prediction is only trustable when the input data follows the distribution of the training dataset. 
When the input data is far away from the training data, the neural networks are inclined to make arbitrarily over-confident predictions~\cite{nguyen2015deep,thulasidasan2019mixup}, hindering the reliability of the deep model in realistic applications.

To tackle this problem, many techniques~\cite{hendrycks2016baseline,hsu2020generalized,liu2020energy,lee2018simple,techapanurak2020hyperparameter,ren2021simple,huang2021importance,liang2017enhancing} have been proposed to facilitate the trained models to recognize those irregular input data and reject to provide predictions.
Specifically, the training data distribution is modelled as the in-distribution, while those irregular data are assumed from the out-of-distribution (OOD). It is the primary goal of differentiating the two types of data, i.e., detecting the out-of-distribution data.
So far, in the literature, OOD samples are typically detected by distinguishing irregular patterns existed in the feature space~\cite{lee2018simple,techapanurak2020hyperparameter,ren2021simple}, logit space~\cite{hendrycks2016baseline,liang2017enhancing}, gradient space~\cite{huang2021importance}, etc. For example, maximum softmax probability~\cite{hendrycks2016baseline} and Helmholtz free energy~\cite{liu2020energy} methods are based on the predicted logits and Mahalanobis distance OOD detector~\cite{lee2018simple} is based on the extracted features.

For the current OOD detection benchmarks, the test dataset is typically a mixture of test data from the in-distribution and some data from a distinct dataset~\cite{lee2018simple,hendrycks2016baseline,liang2017enhancing,hsu2020generalized,lee2017training}.
\begin{figure*}[t]
\centering
\includegraphics[width=0.95\textwidth]{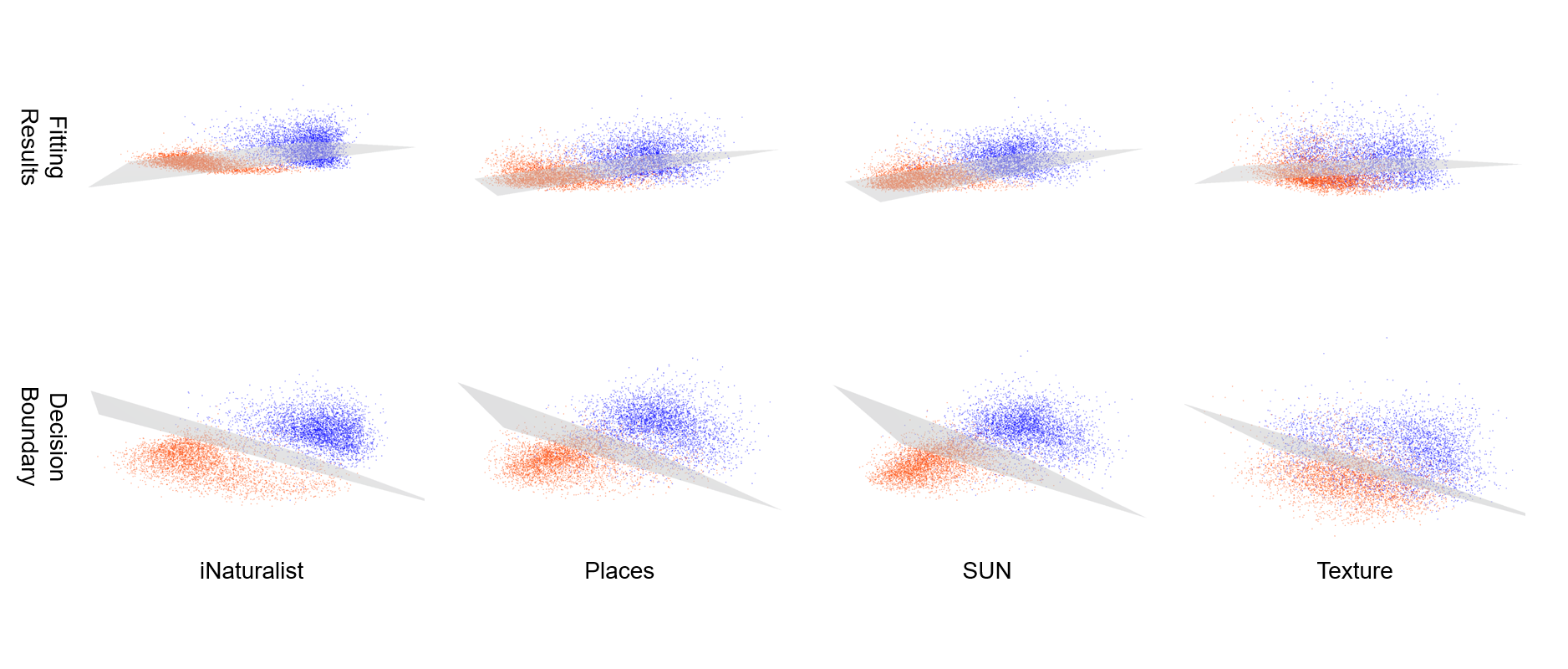}
\vspace{-0.35in}
\caption{
Illustration of the linear relationship between features and OOD scores (EBM energy) of ImageNet(in-distribution) and four OOD datasets. \textit{Blue/orange} points of denote  \textit{in/out-of-distribution} features.
Features were extracted using pre-trained ResNet-101 and reduced to dimension $2$ by PCA for visualization.  Critically, the image was visualized in 3D space(not using 2D points).
We first draw a 3d scatter plot with the reduced feature $(x,y)$ and OOD score $z$ as coordinates, then fit a linear regression and a SVM(assume access to the ground truth labels) via the grey hyper-plane. 
The upper row fits a linear regression model and display the fitted OOD score via the grey hyper-plane, indicating a roughly linear relationship between OOD scores and features.
The lower row presents the fitted decision boundary between in- and out-of-distribution data, suggesting a linear separable feature patterns.
}
\label{fig:linear-separable}
\end{figure*}
In this paper, we show that under such a setup the in-distribution and out-of-distribution data is actually nearly linear-separable as shown in Fig.~\ref{fig:linear-separable}. 
It can be seen that the in-distribution data of ImageNet (blue dots) are roughly linearly separable from the out-of-distribution data (orange dots).

This linear separability is consistently observed across different OOD datasets.

More interestingly, Figure~\ref{fig:linear-separable} conveys that there is an approximated linear relation, which may be learned directly, between the features extracted by the pretrained neural network and the predicted OOD scores generated by different OOD algorithms.
We find then a simple linear regression learned to fit the extracted features and inferred OOD scores can yield more accurate OOD scores than the used base OOD detection algorithms.

Based on this observation, we propose an Embarrassingly simple Test-Time Linear Training (ETLT) model to tackle OOD detection problem at test time.
We model the linear relation between features and OOD scores and provide the rectified OOD score directly based on the input feature.
Specifically, we present two variants of ETLT to handle different scenarios: 
When the feature-score pairs provide strong enough linear relation, a simple Direct Linear Regression (DLR) algorithm is used to learn and produce revised OOD scores.
When the linear relation is less recognisable due to some detection errors introduced by the base OOD detection algorithms, we introduce a Robust Linear Regression (RLR) model to discover the true linear relation from those noisy pairs. Moreover,  DLR can be easily 
implemented as an online version, which is more practical to real-world applications.
We conduct extensive experiments and show the effectiveness of our ETLT across different datasets and over different base OOD detection algorithms.

Our contributions are as follows:
\begin{itemize}
\item We show that there exists a roughly linear separation between the in- and out-of-distribution data according to the existing OOD detection evaluation setup. And we observe a linear relation between the extracted features and the OOD scores inferred by those popular OOD detection methods.
\item We then propose an embarrassingly simple test-time linear training model, including two variants, a simple Direct Linear Regression (DLR) and a Robust Linear Regression (RLR), to discover and use this linear relation to provide revised OOD scores directly based on the input features. We further derive an online version of DLR, which is more practical and produce as good results as  DLR.
\item Our ETLT algorithm enjoys the state-of-the-art performance on various benchmark datasets and improves over four popular base OOD detection methods.
\end{itemize}

\section{Related Work}
\textbf{Out-Of-Distribution Detection}
Analogous to detecting misclassified examples, a baseline of OOD detection was proposed in \cite{hendrycks2016baseline} by using the maximum softmax probability (MSP) produced by the trained neural networks. Inspired by the idea of adversarial
examples~\cite{goodfellow2014explaining}, an input preprocessing method named ODIN~\cite{liang2017enhancing} was proposed to improve MSP. Furthermore, Mahalanobis distance based class-conditional confidence score for OOD detection was proposed in~\cite{lee2018simple} under the modeling of Gaussian discriminant analysis. Though achieving great improvements over MSP, the followup ODIN and Mahalanobis detectors need hyperparameter optimization on a validation set or in-distribution set, which is prohibitive in many cases. To overcome such a drawback, Energy-based score serving as a parameter-free OOD detector was proposed in~\cite{liu2020energy}. Recently, GradNorm was proposed in~\cite{huang2021importance} by exploring the gradient information of minimizing the KL discrepancy between the predicted posterior and the uniform distribution. All the methods above can be applied on top of any pre-trained classification network trained with a cross entropy loss.

Another class of OOD detectors attack on OOD problem by training specifically designed networks~\cite{devries2018learning,hsu2020generalized}. A confidence estimation branch, in addition to the task branch, was proposed in \cite{devries2018learning}. Similarly, Generalized ODIN~\cite{hsu2020generalized} learned separate branches for modeling the joint class-domain probability and domain probability respectively as a decomposition of the confidence scoring, while in this way the temperature scaling could be automatically learned.

Additionally, Outlier Exposure(OE)~\cite{hendrycks2018deep} leveraged an auxiliary OOD dataset to train an OOD detector directly. Out-of-Distribution Mining (ODM)~\cite{masana2018metric} proposed to use metric learning to get rid of the (trouble-maker) softmax layer. ATOM~\cite{chen2021atom} applied adversarial training to shrink the decision boundary between in-distribution dataset and auxiliary OOD dataset to achieve better results. Recently, self-supervised learning had been combined with outlier exposure~\cite{mohseni2020self}.
Furthermore, directly controlling the difference between in- and out-of-distribution samples was proved to further separate them \cite{papadopoulos2021outlier}.

\noindent\textbf{Test-Time Adaptation  (TTA)}.
TTA~\cite{sun2020test,wang2020tent,iwasawa2021test} was introduced to alleviate the distributional shift problem between the training and testing data, as deep models are known to be biased to the training data distribution and could fail predicting correctly on the test data from an unseen distribution. 
TTA adapts the trained models to the novel data during testing by updating its parameters with the mini-batch~\cite{sun2020test,wang2020tent} or full~\cite{iwasawa2021test} unlabeled test data. TTA approaches chose to update batch-norm parameters/statistics to fit the testing data~\cite{wang2020tent,hu2021mixnorm}, or minimize the prediction inconsistency~\cite{zhang2021memo} between different data augmentation of a single data point.
Our ETLT conceptually follows the TTA setup in the way that we have access to the mini-batch or full test data.
However, we focus on using the inferred OOD signal at test and the linear relation between features and OOD scores to improve OOD detection, rather than model adaptation in the standard TTA. The difference between TTA and Transductive Learning(TL) is that TL is learned on  train and test data jointly, while the model for TTA is updated by unlabeled test data at test time as in \cite{wang2020tent}

\noindent\textbf{Outlier Detection}.
Irregular data are also considered in some other research fields.
Outlier detection~\cite{pang2021deep} assumes that the data are polluted by outliers. Generally, OOD detectors process a single sample at a time while outlier detectors assume the accessibility to all the testing samples. When evaluating OOD detection in the TTA setup, the two settings get similar. Therefore, we also compare our algorithms with some outlier detection methods. 
\section{Methodology}
\textbf{Problem setup.}
In the traditional supervised image classification tasks, one typically learns a mapping $f:\mathcal{X}\rightarrow\mathcal{Y}$ from image space $\mathcal{X}\subseteq\mathbb{R}^m$ to label space $\mathcal{Y} = \{1,2,\cdots, C\}$ with a given training set $D = \{(\textbf{x}_i,y_i)\}_{i=1}^n$.
{Then at inference, the predicted label can be achieved according to the maximum score $\hat{y} =\argmax f\left(\textbf{x}\right).$}

In practice, there exists `unexpected' data that is far away from the training set, and for the sake of model reliability those data samples should be rejected without obtaining any predictions.
Formally, we assume that the training data is drawn from the distribution $P_{in}$, which we call in-distribution, and those unexpected data are drawn from a distinct out-distribution $P_{out}$.
Precisely, we assume a mixture distribution $P_{mix}$ defined on $\mathcal{X}\times\mathcal{Z}$, where $\mathcal{Z}=\{0,1\}$, $P_{mix}\left(X|Z=0\right)= P_{in}\left(X\right)$ and $ P_{mix}\left(X|Z=1\right)= P_{out}\left(X\right)$. Out-of-distribution detection aims at distinguishing which distribution $\textbf{x}\sim P_{mix}$ is drawn from.

In general, the OOD detection problem can be viewed as a binary classification problem with only in-distribution data available during training.
However, such a binary decision is hard to achieve in practice. Instead, the continuous OOD scores are usually exploited for OOD detection in the way that the model could reject unreliable instances if their scores are below a proper threshold.
Specifically, for each $\textbf{x}$ we provide an OOD score estimator $S\left(\textbf{x}\right)$ and design a binary classifier with manually-defined threshold $\gamma$:
\begin{equation}
    g(\mathbf{x})= \begin{cases}\text { in, } & \text { if } S(\mathbf{x}) \geq \gamma, \\ \text { out, } & \text { if } S(\mathbf{x})<\gamma.\end{cases}
\end{equation}

Here we review several recent \textbf{OOD score} based methods:

\noindent \textbf{MSP}. The maximum softmax probability~\cite{hendrycks2016baseline} is defined as OOD score given by a trained network
\begin{equation}
S_{\mathrm{MSP}}(\boldsymbol{x}) \coloneqq= \frac{\exp \left( f_{\hat{y}}(\boldsymbol{x})/T \right)}{\sum_{j=1}^C{\exp}\left( f_j(\boldsymbol{x})/T \right)},
\end{equation}
where $T$ is the temperature coefficient introduced in~\cite{liang2017enhancing} to make the softmax prediction sharp. Vanilla MSP~\cite{hendrycks2016baseline} did not include temperature scaling, which was proposed in ODIN~\cite{liang2017enhancing}. For the ease of our overall formulation we add it here.

\noindent \textbf{Energy}. Energy-based model (EBM)~\cite{lecun2006tutorial} aims to find a suitable energy function $E\left(x,y\right)$ defined on $\mathcal{X}\times\mathcal{Y}$, and model the posterior probabilty by the Gibbs distribution
\begin{equation}
p(y \mid \mathbf{x})=\frac{\exp(-E(\mathbf{x}, y) / T)}{\exp(-E(\mathbf{x}) / T)}.
\end{equation}
The Helmholtz free energy $E\left(\mathbf{x}\right)$ of a given data point $\mathbf{x}$ can be expressed as the negative of the log partition function
\begin{equation}
E(\mathbf{x})=-T \cdot \log \int_{y^{\prime}} \exp(-E\left(\mathbf{x}, y^{\prime}\right) / T).
\end{equation}
The Helmholtz free energy for $\mathbf{x}\sim P_{in}$ is push down during the training process, and therefore can serve as an alternative metric for OOD detection~\cite{liu2020energy}. More details of EBM can be found in ~\cite{lecun2006tutorial}. Direct connecting Helmholtz free energy with softmax probability, we get another baseline OOD detector:
\begin{equation}
S_{energy}(\mathbf{x}) \coloneqq - E(\mathbf{x}) = T \cdot \log \sum_{i}^{C} \exp(f_{i}(\mathbf{x}) / T).
\end{equation}

\noindent \textbf{KL}. An in-distribution sample should tend to have a prediction with the minimum entropy, while an OOD data is not properly defined in the training label space and should be predicted with large predictive entropy. Therefore, the Kullback-Leibler (KL) divergence between the softmax output and a uniform distribution is proposed in \cite{lee2017training} to improve the OOD scoring
\begin{equation}
\begin{aligned}
S_{\mathrm{KL}}(\mathbf{x}) &\coloneqq
 D_{\mathrm{KL}}\left(\mathbf{u} \| \mathrm{SoftMax}(f(\boldsymbol{x}))\right)\\
 &=-\frac{1}{C} \sum_{i=1}^{C} \log \frac{\exp(f_{i}(\mathbf{x}) / T)}{\sum_{j=1}^{C} \exp(f_{j}(\mathbf{x}) / T)}-H(\mathbf{u})
\end{aligned}
\end{equation}

\noindent where $\mathbf{u}=[1 / C, 1 / C, \ldots, 1 / C] \in \mathbb{R}^{C}$ is a uniform probability and $H(\mathbf{u}) = \log C$. Since we focus on the OOD score, thus the constant term could be dropped. 

\noindent \textbf{ODIN}. ODIN~\cite{liang2017enhancing} got the inspiration from adversarial attack~\cite{goodfellow2014explaining} and find that including the adversarial perturbed inputs into training improves the final OOD scoring.
Given an input image $\boldsymbol{x}$ and the predicted softmax probability, ODIN first generates the perturbed image by
\begin{equation}
    \tilde{\boldsymbol{x}}=\boldsymbol{x}-\varepsilon \operatorname{sign}\left(-\nabla_{\boldsymbol{x}} \log S_{\mathrm{MSP}}(\boldsymbol{x})\right),
\end{equation}
then the MSP of the perturbed image will be maximized during training.
At test time, the perturbed image $\tilde{\boldsymbol{x}}$ is used to produce the final OOD score as
\begin{equation}
    S_{\mathrm{ODIN}}(\boldsymbol{x})=S_{\mathrm{MSP}}(\tilde{\boldsymbol{x}}).
\end{equation}
In the following, we will demonstrate a linear relation between the extracted features and inferred scores at test time and learn a linear regression to improve the above four base OOD detection methods.
\subsection{The Surprisingly Linear Relation}
We think that the above four OOD scores are likely to be sub-optimal, because they only deal with a single sample at a time and ignore the interaction between samples. 

In this paper, we show that with a mini-batch (full) test data in hand, we can improve the baseline OOD methods significantly. 
Although there is a certain number of wrongly inferred OOD scores of the test data, i.e. the OOD samples over-confidently predicted by the trained model, training a linear regression is still able to rectify those overconfident (noisy) scores.

We empirically observe that a simple linear relationship exists between the feature extracted from the neural network and the inferred OOD scores by those base OOD detection algorithms, as shown in Fig.~\ref{fig:linear-separable}. Furthermore, the in- and out-of-distribution features are linear-separable. We also give a 2D visualization in Fig.~\ref{fig:ImageNet-Separability}.
Although those OOD scores are from different (nonlinear) scoring algorithms, the input features and OOD scores are well fitted with a linear regression. 

We thought this roughly linear  relation is quite intuitive, as it is just a direct linear approximation to  Generalized Linear Model (GLM).
Particularly, the typical OOD methods such as MSP has a backbone and softmax-based classification header and an OOD scorer. As softmax function is a typical GLM, the relationship of softmax-based  classification head and OOD score can be taken as roughly GLM.
During  training stage, the model should be trained to match the classifier head and feature backbone. Therefore,  the linear relation between feature and OOD score is established as a consequence of an approximation of the GLM.
Based on this empirical observation, we propose a simple method, which enjoys nearly no extra cost, to improve the current OOD detection methods.

\subsection{ Simple Test-Time Linear Training}
Let us denote the (un-normalized) OOD score of an input image as $s\coloneqq S(\mathbf{x})$.
We assume a linear relation between the OOD score $s$ and the input feature $\mathbf{z}$ extracted by the trained model:
\begin{equation}
s=\mathbf{z}^{\top}\beta+\varepsilon.
\end{equation}
For a particular OOD algorithm $\omega$, it will produce a (noisy) estimation $\hat{s}_i$ of $s_i$ based on the input image $\mathbf{x}_i$, resulting $\hat{s}_i\coloneqq S_{\omega}(\mathbf{x}_i)$.
We aim to discover the \emph{true} $\beta$ from the feature-score pair $(\mathbf{z}_i,\hat{s}_i)$, hence we can get a more precise estimation of $s_i$.

We consider two slightly different test-time linear training methods. Particularly, when a set of test instances are embedded with a small amount of noisy scores, a simple linear regression model could be well fitted.

On the other hand, if the amount of noisy samples is too large to train a linear model, we will introduce a robust variant which generates the noise scores such that a clean set could be selected. 
The details of two models are introduced as follows.

\noindent \textbf{Direct Linear Regression (DLR)}.
When linear relation is recognisable, a simple linear regression model is sufficient to estimate the true $\beta$ such that:
\begin{equation}
\hat{\beta}=\argmin_{\beta} \sum_{i=1}^n (\hat{s}_i-\mathbf{z}^{\top}_i\beta)^2,
\end{equation}
which yields the closed-form solution 
\begin{equation}
\hat{\beta}=(\mathbf{Z}^{\top}\mathbf{Z})^{\dagger}\mathbf{Z}^{\top}\hat{\mathbf{S}},
\end{equation}
where $\mathbf{Z}$ and $\hat{\mathbf{S}}$ are the stack of $\mathbf{z}_i$ and $\hat{s}_i$ by rows, respectively.
With this estimator, we could directly provide our OOD estimator for instance $i$ as:
\begin{equation}
\hat{s}_{\mathrm{ours}}=\mathbf{z}^{\top}_i\hat{\beta}.
\end{equation}

\noindent \textbf{Robust Linear Regression (RLR)}.
When the originally inferred OOD scores are too noisy caused by the OOD scoring method $\omega$, we can design another outlier detection model to remove those noisy data from estimating $\beta$.

Specifically, we introduce an explicit data-dependent variable $\gamma_i$ to represent the prediction error of the linear regression model, such that 
\begin{equation}
\hat{s}_i=\mathbf{z}_i^{\top}\beta+\gamma_i+\varepsilon.
\end{equation}
When $|\gamma_i|$ is small, the linear relation is preserved, indicating that instance $i$ is helpful to guide the estimation of $\beta$.
On the contrary, when $|\gamma_i|$ is large, the corresponding instance should be ignored as the estimated $\hat{s}_i$ is likely to be wrong.

Note that in this robust variant, we aim to detect the noisy instances instead of predicting $\beta$, hence we only care about the solution of $\gamma_i$.
Based on this intuition, we design the following optimization problem:
\begin{equation}
\min_{\beta,\gamma} \sum_{i=1}^n\left[ \frac{1}{2}(\hat{s}_i-\mathbf{z}^{\top}_i\beta-\gamma_i)^2 + \lambda |\gamma_i|\right].
\end{equation}
When all $\gamma_i$ are resolved, we can directly get the closed-form estimation of $\beta$ as $\hat{\beta}=(\mathbf{Z}^{\top}\mathbf{Z})^{\dagger}\mathbf{Z}^{\top}(\hat{\mathbf{S}}-\gamma)$.
Substitute it into the objective and further define that $\tilde{\mathbf{Z}}=\mathbf{I}-\mathbf{Z}(\mathbf{Z}^{\top}\mathbf{Z})^{\dagger}\mathbf{Z}^{\top}$ and $\tilde{\mathbf{S}}=\tilde{\mathbf{Z}}\hat{\mathbf{S}}$,
we can simplify the objective as
\begin{equation}
\min_{\gamma} \frac{1}{2}\Vert \tilde{\mathbf{S}}-\tilde{\mathbf{Z}}\gamma\Vert_2^2+ \lambda \Vert\gamma\Vert_1,
\end{equation}
which is a robust linear regression problem for $\gamma$.
We can then select a proper $\lambda$ to solve $\gamma$ and select the most reliable subset to estimate $\beta$ as \begin{equation}
\hat{\beta}=(\mathbf{Z}_{\mathrm{sub}}^{\top}\mathbf{Z}_{\mathrm{sub}})^{\dagger}\mathbf{Z}_{\mathrm{sub}}^{\top}\hat{\mathbf{S}}_{\mathrm{sub}}.
\end{equation}

\begin{algorithm}[tb]
   \caption{ETLT}
   \label{alg:linear revision}
   {\bfseries Input:} image $\boldsymbol{x}_i,1\le i\le n$, an  OOD score function $S:\mathcal{X}\rightarrow \mathbb{R}$ 
   
   Calculate the features of all images $\mathbf{z}_i = g\left(\boldsymbol{x}_i\right)$
   
  Calculate the OOD score of all images $\hat{s}_i = S\left(\boldsymbol{x}_i\right)$
  
  Apply DLR to use the whole training set, or RLR to get a subset.
  
    $\hat{\beta}=\argmin_{ \beta}\sum_{i\in\Gamma}^n{\left( \hat{s}_i-\mathbf{z}_i^{T}\beta  \right) ^2}$
   
   Compute calibrated OOD score $s_i=\mathbf{z}_i^{T}\hat{\beta}$
   
   Using $s_i$ as the OOD score.

\end{algorithm}

\begin{algorithm}[tb]
   \caption{RLR}
   \label{alg:RLR}

   {\bfseries Input:} features $\mathbf{z}_i$ and OOD score $\hat{s}_i$, $1\le i\le n$,
   
   Normalize $\mathbf{z}_i$ to unit Euclidean norm
    
   Apply dimensionality reduction on $\mathbf{z}_i$ to $d\ll n$
   
   Stack $\mathbf{z}_i$ and $\hat{s}_i$ by rows to $\mathbf{Z}$ and $\hat{\mathbf{S}}$ 
   
  Calculate projection $\tilde{\mathbf{Z}}=\mathbf{I}-\mathbf{Z}(\mathbf{Z}^{\top}\mathbf{Z})^{\dagger}\mathbf{Z}^{\top}$ and $\tilde{\mathbf{S}}=\tilde{\mathbf{Z}}\hat{\mathbf{S}}$
  
   Solving Lasso $\hat{\gamma} = \argmin_{\gamma} \frac{1}{2}\Vert \tilde{\mathbf{S}}-\tilde{\mathbf{Z}}\gamma\Vert_2^2+ \lambda \Vert\gamma\Vert_1$
   
   Select a subset $\hat{\mathbf{Z}}$ with the lowest $p\%$ of $|\hat{\gamma}_i|$.
   
   return $\hat{\mathbf{Z}}$

\end{algorithm}

\subsection{Online OOD detection}
In the above methodology, we assume the availability of the full test set or large batch of test instances. One may judge that the test set could not be accessed fully. Therefore, we also provide a online version as per those test-time adaptation works~\cite{wang2020tent,sun2020test,iwasawa2021test}.
The test data comes in a stream batch by batch, thus we derive a batch-wise version of our ETLT, which updates the linear model with the current and past mini-batch data iteratively. Here we illustrate the wisdom of how to estimate $\beta$ online. Let us assume we have two mini-batch data pairs $(\mathbf{Z}_1, \hat{\mathbf{S}}_1)$, $(\mathbf{Z}_2, \hat{\mathbf{S}}_2)$. Then, we can obtain two block matrices,
$$
\mathbf{Z}=\left[ \begin{array}{c}
	\mathbf{Z}_1\\
	\mathbf{Z}_2\\
\end{array} \right] , S=\left[ \begin{array}{c}
	\hat{\mathbf{S}}_1\\
	\hat{\mathbf{S}}_2\\
\end{array} \right].
$$

Recall that $\hat{\beta}=(\mathbf{Z}^\top\mathbf{Z})^{\dagger}\mathbf{Z}^{\top}\hat{\mathbf{S}}$, then we can get
\begin{equation}
    \mathbf{Z}^{\top}\mathbf{Z}\,\,=\,\,\left[ \begin{matrix}
	\mathbf{Z}_{1}^{\top}&		\mathbf{Z}_{2}^{\top}\\
\end{matrix} \right] \left[ \begin{array}{c}
	\mathbf{Z}_1\\
	\mathbf{Z}_2\\
\end{array} \right] =\mathbf{Z}^{\top}_1\mathbf{Z}_1+\mathbf{Z}^{\top}_2\mathbf{Z}_2,
\end{equation}
\begin{equation}
    \mathbf{Z}^{\top}\hat{\mathbf{S}}\,\,=\,\,\left[ \begin{matrix}
	\mathbf{Z}_{1}^{\top}&		\mathbf{Z}_{2}^{\top}\\
\end{matrix} \right] \left[ \begin{array}{c}
	\hat{\mathbf{S}}_1\\
	\hat{\mathbf{S}}_2\\
\end{array} \right] =\mathbf{Z}^{\top}_1\hat{\mathbf{S}}_1+\mathbf{Z}^{\top}_2\hat{\mathbf{S}}_2.
\end{equation}
When more mini-batch data pairs $(\mathbf{Z}_i, \hat{\mathbf{S}}_i)$ come, we thus can update $\mathbf{Z}^{\top}\mathbf{Z}$ and $\mathbf{Z}^{\top}\hat{\mathbf{S}}$ according to Algorithm.~\ref{alg:online linear revision}

\begin{algorithm}[t]
   \caption{Online DLR}
   \label{alg:online linear revision}
   {\bfseries Input:} an  OOD score function $S:\mathcal{X}\rightarrow \mathbb{R}$, batch size $b$, numbers of batch $m$
   
   {\bf Initialize:}  $\textbf{A}\leftarrow0$,  $\textbf{b}\leftarrow0$
   
   \For{$k = 1,\cdots,m$}
   {Sample $b$ images $\boldsymbol{x}_i, 1\le i\le b$
   
   Calculate the features of all images $\mathbf{z}_i = g\left(\boldsymbol{x}_i\right)$
   
   Calculate the OOD score of all images $\hat{s}_i = S\left(\boldsymbol{x}_i\right)$
   
    Stack $\mathbf{z}_i$ and $\hat{s}_i$ by rows to $\mathbf{Z}$ and $\hat{\mathbf{S}}$ 
   
   $\textbf{A}\leftarrow\textbf{A}+\mathbf{Z}^{\top}\mathbf{Z}$
   
   $\textbf{b}\leftarrow\textbf{b}+\mathbf{Z}^{\top}\hat{\mathbf{S}}$
   
   $\hat{\beta}=(\textbf{A})^{\dagger}\textbf{b}$
   
   Compute calibrated OOD score of the present batch $s_i=\mathbf{z}_i^{T}\hat{\beta}$
   }
   
   Using $s_i$ as the OOD score.
\end{algorithm}

\begin{figure}
\centering
\begin{subfigure}
\centering
\includegraphics[width=0.35\linewidth]{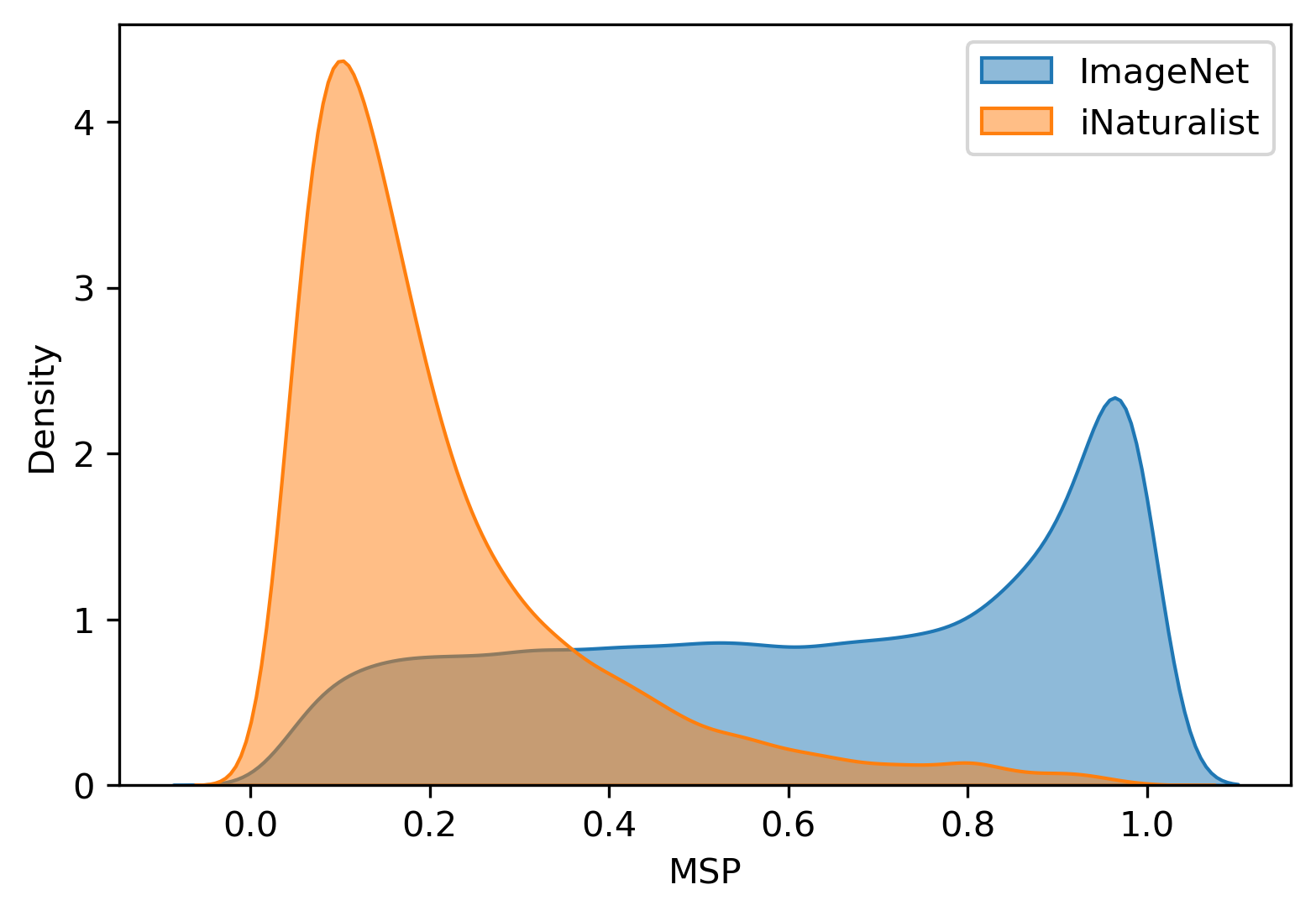}
\end{subfigure}
\begin{subfigure}
\centering
\includegraphics[width=0.35\linewidth]{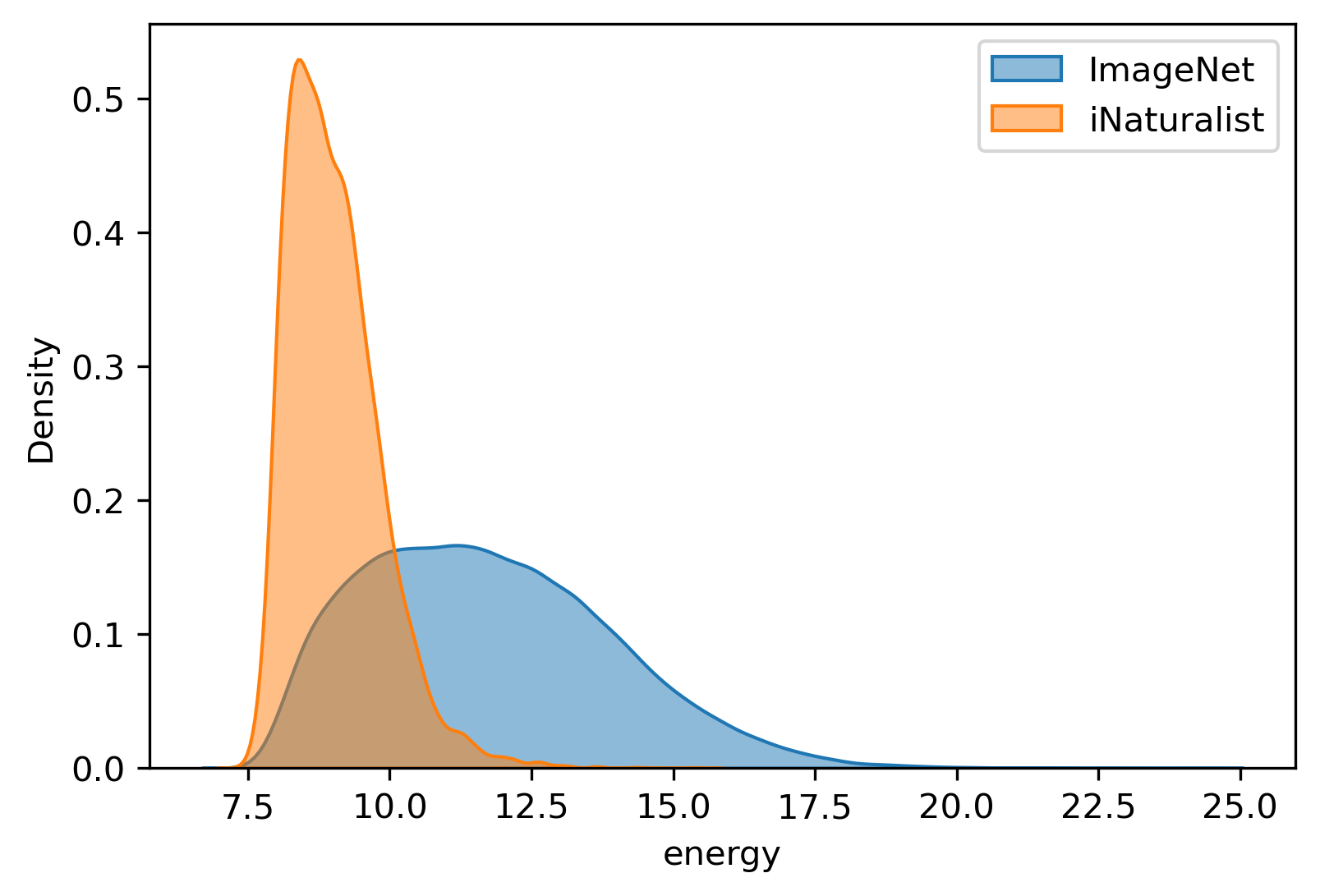}
\end{subfigure}
\vspace{-0.1in}
\caption{\label{Distribution:ImageNet}Kernel density estimate plot of the  in- and out-of-distribution samples. }
\vspace{-0.15in}
\end{figure}

\subsection{The Linear Separability}
\label{sec:linear separability}
\begin{figure}
\label{LinearSeparation:ImageNet}
\centering
\begin{subfigure}
\centering
\includegraphics[width=0.35\linewidth]{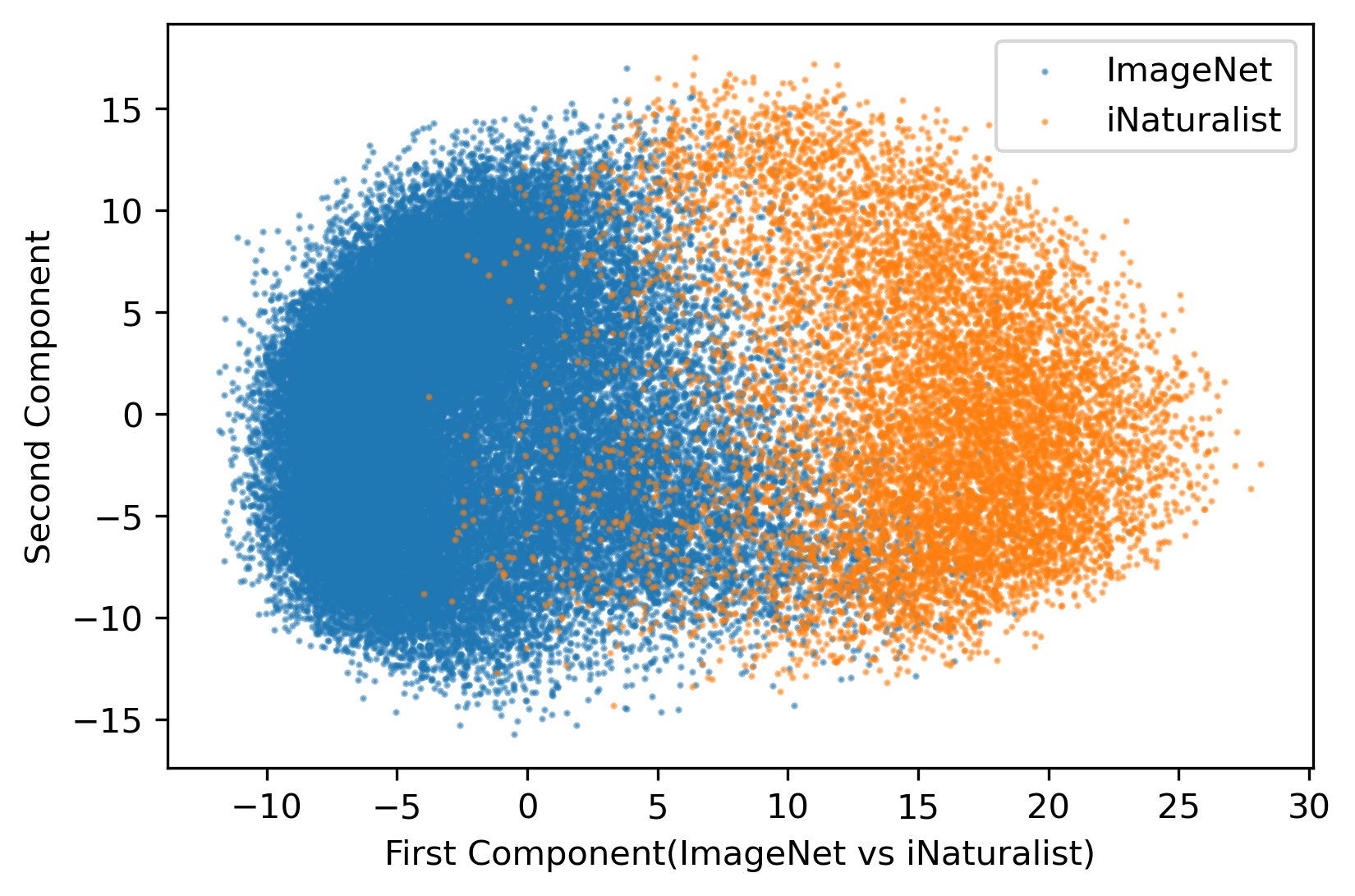}
\end{subfigure}
\begin{subfigure}
\centering
\includegraphics[width=0.35\linewidth]{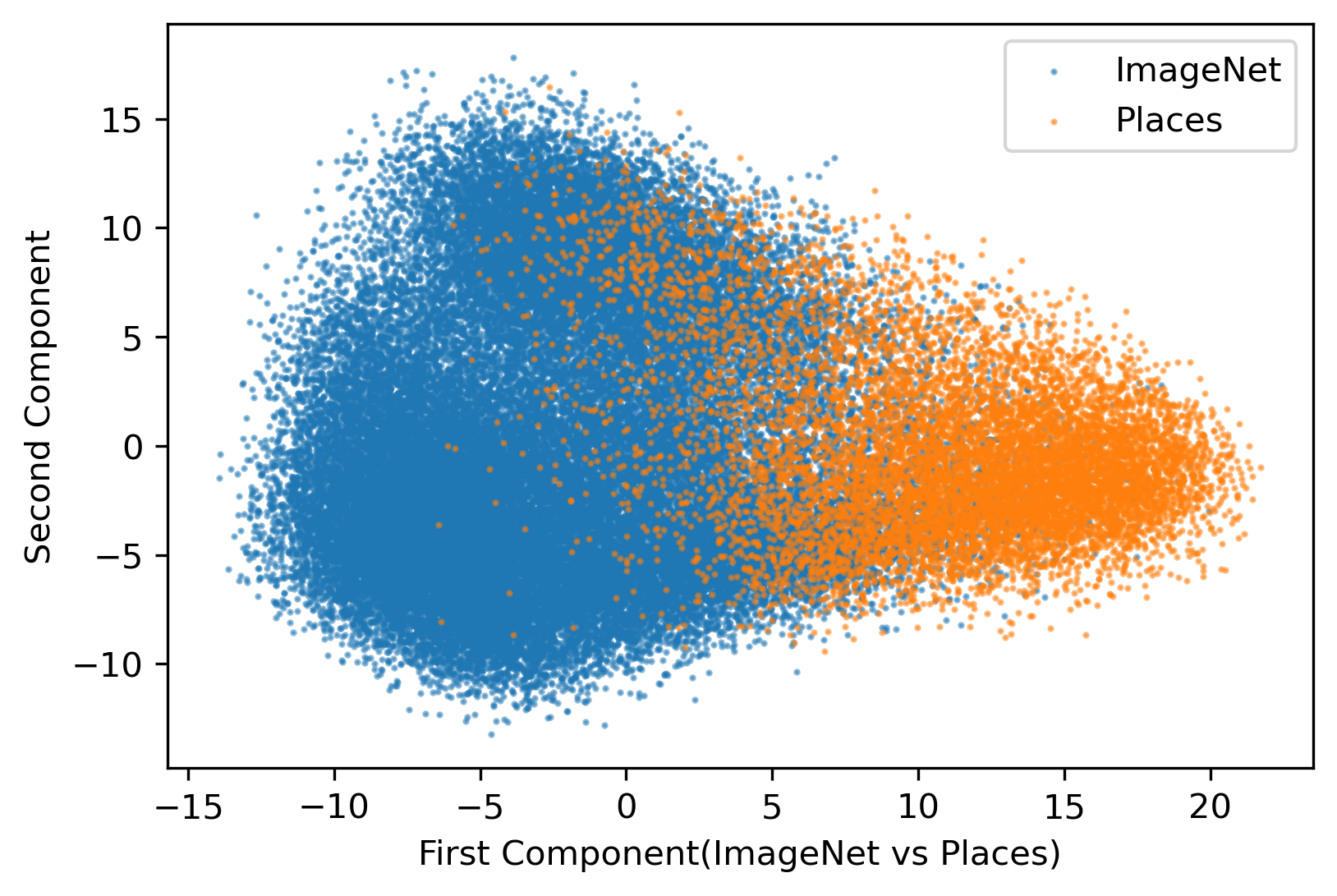}
\end{subfigure}
\vspace{-0.1in}
\caption{Visualization of features from ImageNet and two OOD datasets. The dimension of features is reduced to two by PCA.  ~\label{fig:ImageNet-Separability}}
\vspace{-0.25in}
\end{figure}

Given an in-distribution dataset, and we sample the OOD data from a specific dataset. We empirically show that the two types of data tend to be linearly separated in Section~\ref{sec:intro}.
As we visualize in Fig~\ref{fig:ImageNet-Separability}, a simple unsupervised feature reduction by PCA can cluster the in-distribution and OOD data well, indicating that the feature patterns are good enough to be distinguished.
Furthermore, as shown in Fig.~\ref{fig:linear-separable}, a hyper-plane can be roughly approximated to model the linear relation between the features and OOD scores.
Intuitively, the linear separation comes from the definition of OOD data which has a different distribution from  training dataset combined with the powerful feature extractor of neural network.

One potential issue with such linear relation modeling is that in our visualization, those feature points are not strictly on the hyper-plane.
Instead, most of them are around the hyper-plane.
We argue that this inconsistency is from the noisy and biased estimation given by OOD detection algorithms, otherwise no score rectification is needed.
To show this,
we draw OOD scores of in-distribution  and out-of-distribution data using popular baseline OOD detection algorithms as  in Fig.~\ref{Distribution:ImageNet}.
This suggests that the OOD score distributions are not disjoint between in-distribution and out-of-distribution data.
Hence they cannot produce a perfect OOD detection.
However, we will show that these baseline OOD detector, although not powerful enough, can provide necessary information for distinguishing in- and out-of distribution data.
We show that with the near linear-separable features and OOD scores, our proposed ETLT can provide a more precise OOD scoring and thus improve OOD detection, usually by a large margin. 
\label{sec:linear-relation}

\subsection{Small Scale OOD experiments}
\label{section:cifar}

We conduct experiments on the  CIFAR-10 and CIFAR-100 datasets. Although CIFAR datasets contain easy classification images, it is hard for OOD detection due to its low resolution, especially on CIFAR-100. We use a Wide ResNet~\cite{zagoruyko2016wide} of 40 layers trained on the training sets of CIFAR-10 and CIFAR-100. We use the CIFAR test set as in-distribution data and sample 2000 images from six different out-of-distribution datasets including Textures~\cite{cimpoi2014describing}, SVHN~\cite{netzer2011reading},
Places365~\cite{zhou2017places}, LSUN-Crop~\cite{yu2015lsun}, LSUN-Resize~\cite{yu2015lsun}, and iSUN~\cite{xu2015turkergaze} following the setting in~\cite{liu2020energy}. Due to randomness, we repeat 10 times trials and report the average results. We set temperature $T=1$ for all classifiers. For ODIN,  noise level is set to $\epsilon = 0.0024$. We set $p=80$ in Algorithm~\ref{alg:RLR}. Besides those popular base OOD detectors, we also compare our algorithm with a newly proposed OOD detector GradNorm~\cite{huang2021importance} and three traditional methods: Gaussian Mixture Models (GMM), Local Outlier Factor (LOF) and Isolation Forest (IF).

The results are shown in Table~\ref{tabel:CIFAR},  where results show that our test time ETLT can improve the base OOD detection methods in almost all cases. Moreover, results in Table~\ref{tabel:CIFAR} show that our robust method RLR can boost the performance further. Our RLR outperforms DLR by $7.9\%$ FPR95 and $2.56\%$ AUROC on CIFAR-100 when using MSP as our base scoring function. Consistent improvements can be observed when using other scoring functions, such as Energy, ODIN and KL. We can also see that our DLR and RLR outperforms the other transductive methods IF, LOF and GMM and a recent state of art method GradNorm.
It is worth noting that our algorithm perform even competitive with Outlier Exposure~\cite{hendrycks2018deep} and fine-tuned energy score~\cite{liu2020energy}, which utilize an extra outlier dataset to direct train the network to distinguish in- and out-of-distribution data.

\begin{table}[t]
\centering
\caption{Results on CIFAR datasets. The best and second best results are highlighted by the fonts of text bold and italic, respectively. \label{tabel:CIFAR}} 
\begin{tabular}{lcccc}
\toprule
In Dataset & \multicolumn{2}{c}{CIFAR-10} & \multicolumn{2}{c}{CIFAR-100}  \tabularnewline
Metric & FPR95$\downarrow$ & AUROC$\uparrow$ & FPR95$\downarrow$ & AUROC$\uparrow$ \tabularnewline
\midrule
\midrule
Softmax & 51.37 & 90.87 & 80.21 & 75.67  \tabularnewline
+RLR & \textit{13.50} & \textit{96.42} & \textbf{43.73} & \textbf{87.06}\tabularnewline
+DLR & \textbf{12.30} & \textbf{97.01} & \textit{51.63} & \textit{84.50}\tabularnewline\midrule
Energy & 32.98 & 91.88 & 73.46 & 79.67 \tabularnewline
+RLR & 16.14 & 95.64& 58.06 & 82.43\tabularnewline
+DLR & 18.01 & 94.96 & 60.63 & 81.15 \tabularnewline\midrule
ODIN & 35.77 & 90.96 & 74.55 & 77.23\tabularnewline
+RLR & 35.79 & 87.24& 51.73 & 82.74\tabularnewline
+DLR & 37.94 & 86.10 & 51.87 & 81.52 \tabularnewline\midrule
KL & 32.98 & 91.88 & 73.46 & 79.67 \tabularnewline
+RLR & 16.12 & 95.64 & 58.06 & 82.43 \tabularnewline
+DLR & 18.01 & 94.96 & 60.63 & 81.15\tabularnewline\midrule
IF & 79.96 & 62.47  & 80.91 & 66.15 \tabularnewline
LOF & 95.81 & 56.45 & 98.23 & 43.32\tabularnewline
GMM & 87.70 & 58.28& 94.06 & 69.96\tabularnewline
GradNorm & 59.84 & 71.65& 86.55 & 57.56\tabularnewline
\bottomrule
\end{tabular}
\vspace{-0.15in}
\end{table}

\subsection{Large Scale OOD experiments}
\label{section:large}
\begin{table*}

\caption{Results on ImageNet-1k and iNaturalist/SUN/Places/Textures datasets. The best and second best results are highlighted by the bold and italic font, respectively.}
\label{tabel:imagenet-1k}
\centering
\resizebox{0.9\linewidth}{!}{
\begin{tabular}{lcccccccccc}
\toprule
OOD Dataset & \multicolumn{2}{c}{iNaturalist} & \multicolumn{2}{c}{SUN} & \multicolumn{2}{c}{Places} & \multicolumn{2}{c}{Textures} & \multicolumn{2}{c}{Average} \tabularnewline
Metric & FPR95 & AUROC & FPR95 & AUROC & FPR95 & AUROC & FPR95 & AUROC & FPR95$\downarrow$ & AUROC$\uparrow$ \tabularnewline
\midrule
\midrule
IF & 88.58 & 61.60 & 90.12 & 57.85 & 93.45 & 50.24 & \textbf{54.34} & \textbf{87.76} & 81.62 & 64.36 \tabularnewline
LOF & 95.16 & 51.57 & 94.89 & 52.27 & 93.05 & 56.37 & 82.02 & 65.39 & 91.28 & 56.40 \tabularnewline
GMM & 87.90 & 68.43 & 89.99 & 63.29 & 96.85 & 52.83 & 95.37 & 35.34 & 92.53 & 54.97 \tabularnewline
GradNorm & 50.03 & 90.33 & \textbf{46.48} & \textit{89.03} & 60.86 & \textbf{84.82} & 61.42 & \textit{81.07} & 54.70 & \textit{86.31} \tabularnewline
\midrule
MSP & 63.69 & 87.59 & 79.98 & 78.34 & 81.44 & 76.76 & 82.73 & 74.45 & 76.96 & 79.29 \tabularnewline
+RLR & \textit{21.84} & \textit{94.87} & 53.27 & 86.46 & 59.06 & 83.79 & 59.02 & 80.01 & \textit{48.30} & 86.28 \tabularnewline
+DLR & \textbf{21.00} & \textbf{94.98} & \textit{50.68} & 87.13 & \textbf{57.16} & \textit{84.47} & \textit{58.48} & 80.24 & \textbf{46.83} & \textbf{86.71} \tabularnewline
\midrule
Energy & 64.91 & 88.48 & 65.33 & 85.32 & 73.02 & 81.37 & 80.87 & 75.79 & 71.03 & 82.74 \tabularnewline
+RLR & 46.51 & 91.29 & 53.23 & 88.82 & 64.42 & 83.73 & 76.49 & 73.76 & 60.16 & 84.40 \tabularnewline
+DLR & 45.48 & 91.04 & 52.11 & 88.67 & 62.71 & 84.35 & 69.49 & 75.39 & 57.45 & 84.86 \tabularnewline
\midrule
KL & 64.91 & 88.48 & 65.32 & 85.31 & 73.02 & 81.37 & 80.87 & 75.79 & 71.03 & 82.74 \tabularnewline
+RLR & 46.50 & 91.29 & 53.23 & 88.82 & 64.42 & 83.73 & 76.49 & 73.76 & 60.16 & 84.40 \tabularnewline
+DLR & 45.48 & 91.04 & 52.10 & 88.67 & 62.71 & 84.35 & 69.49 & 75.40 & 57.44 & 84.86 \tabularnewline
\midrule
ODIN & 62.69 & 89.36 & 71.67 & 83.92 & 76.27 & 80.67 & 81.31 & 76.30 & 72.99 & 82.56 \tabularnewline
+RLR & 37.28 & 92.49 & 54.51 & 87.50 & 62.87 & 83.48 & 66.95 & 76.36 & 55.40 & 84.96 \tabularnewline
+DLR & 34.84 & 92.94 & 51.31 & \textbf{92.94} & \textit{60.54} & \textit{84.47} & 66.44 & 76.85 & 53.28 & 85.68 \tabularnewline
\bottomrule
\end{tabular}
}
\end{table*}
For large scale OOD detection, we use the ImageNet-1k benchmark following ~\cite{huang2021importance}. We use the validation set of ImageNet-1k as the in-distribution data, which consists of 50000 natural images with 1000 categories.
The out-of-distribution data consist of four datasets, iNaturalist~\cite{van2018inaturalist}, SUN~\cite{xiao2010sun}, Places~\cite{zhou2017places} and Textures~\cite{cimpoi2014describing}.
Google BiT-S models of ResNetV2-101 trained on ImageNet-1k is used as the feature extractor. For MSP, energy score and KL divergence, we set temperature $T = 1$. For ODIN, temperature is set to $T = 1000$, with noise level of $\varepsilon = 0$ since FGSM will not improve the results. For $p$ in RLR, we set $p=80$.

The results are shown in Table ~\ref{tabel:imagenet-1k},  where every OOD set are test separately and the mean results are calculated. The results show that our algorithms achieve consistent improvement over the four baseline OOD detectors. For MSP, our linear revision improve FPR95 from $76.96\%$ to $46.83\%$, about the $30.13\%$ of the improvement and outperform the state-of-the-art OOD score GradNorm by $7.87\%$. For AUROC, the linear calibrated MSP produce the best result, achieving equally good results as GradNorm. It seems on ImageNet-1k, the robust linear regression will not improve the results but result in slight decrease. {We assume this is because the ImageNet-1k model, which was pretrained with sufficient data, owns powerful feature extraction. Thus, on ImageNet-1k OOD benchmarks the linear relation is strong enough to learn directly by DLR without the need of RLR.
}

\begin{table}[t]
\centering
\caption{Results on CIFAR under different OOD detector.  \label{tabel:ChangeOnlineBaselineDetector}} 
\begin{tabular}{lcccc}
\toprule
In Dataset & \multicolumn{2}{c}{CIFAR-10} & \multicolumn{2}{c}{CIFAR-100}  \tabularnewline
Metric Size & FPR95$\downarrow$ & AUROC$\uparrow$ & FPR95$\downarrow$ & AUROC$\uparrow$ \tabularnewline
\midrule
\midrule
MSP & 51.37 & 90.87 & 80.21 & 75.67  \tabularnewline
+Online DLR & 15.06 & 96.05 & 55.84 & 83.31  \tabularnewline \midrule
Energy & 32.98 & 91.88 & 73.46 & 79.67  \tabularnewline
+Online DLR & 18.57 & 94.89 & 62.00 & 80.91  \tabularnewline \midrule
ODIN & 35.77 & 90.96 & 74.55 & 77.23  \tabularnewline
+Online DLR & 40.30 & 85.15 & 54.66 & 80.24  \tabularnewline \midrule
KL & 32.98 & 91.88 & 73.46 & 79.67  \tabularnewline
+Online DLR & 18.57 & 94.89 & 62.00 & 80.91  \tabularnewline 
\bottomrule
\end{tabular}
\end{table}

\subsection{Online Test-Time Training}
In the previous experiments, the full test set is accessed during OOD detection, which is a special case of online test-time learning~\cite{wang2020tent,sun2020test,iwasawa2021test}, i.e. the batch size equals the total number of test data. Here we further check how the performance varies when the batch size gets smaller, i.e. the test time data comes in stream. We compare our online DLR with those popular base OOD methods in this online setup in Table~\ref{tabel:ChangeOnlineBaselineDetector}. From the results, we can see that our online DLR still consistently outperforms the base OOD detection methods, only except ODIN on CIFAR-10, demonstrating the efficacy of our online DLR. More importantly, the results of our online DLR are close to the results in Table~\ref{tabel:CIFAR}, where the full test data set was accessed during model training. These nearly identical results clearly indicates that the effectiveness of our proposed method is not from the access of the full test set, but from the effective transductive learning regression learning as per those TTA methods~\cite{wang2020tent,sun2020test,iwasawa2021test}. 
\subsection{Ablation Study}

\subsubsection{Effect of Subset Selection of RLR}
\begin{figure}
\centering
\begin{subfigure}
\centering
\includegraphics[width=0.35\linewidth]{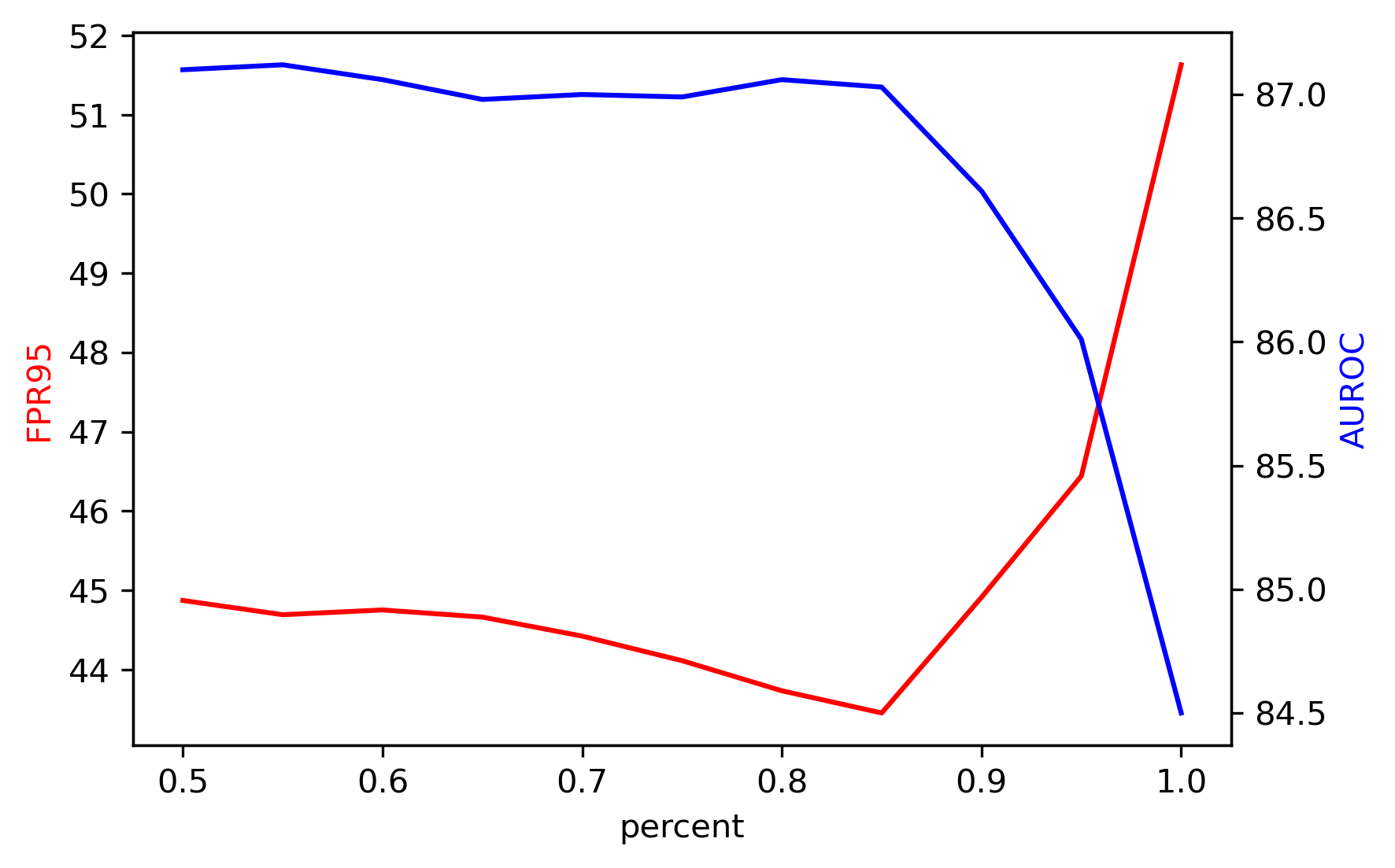}
\end{subfigure}
\begin{subfigure}
\centering
\includegraphics[width=0.35\linewidth]{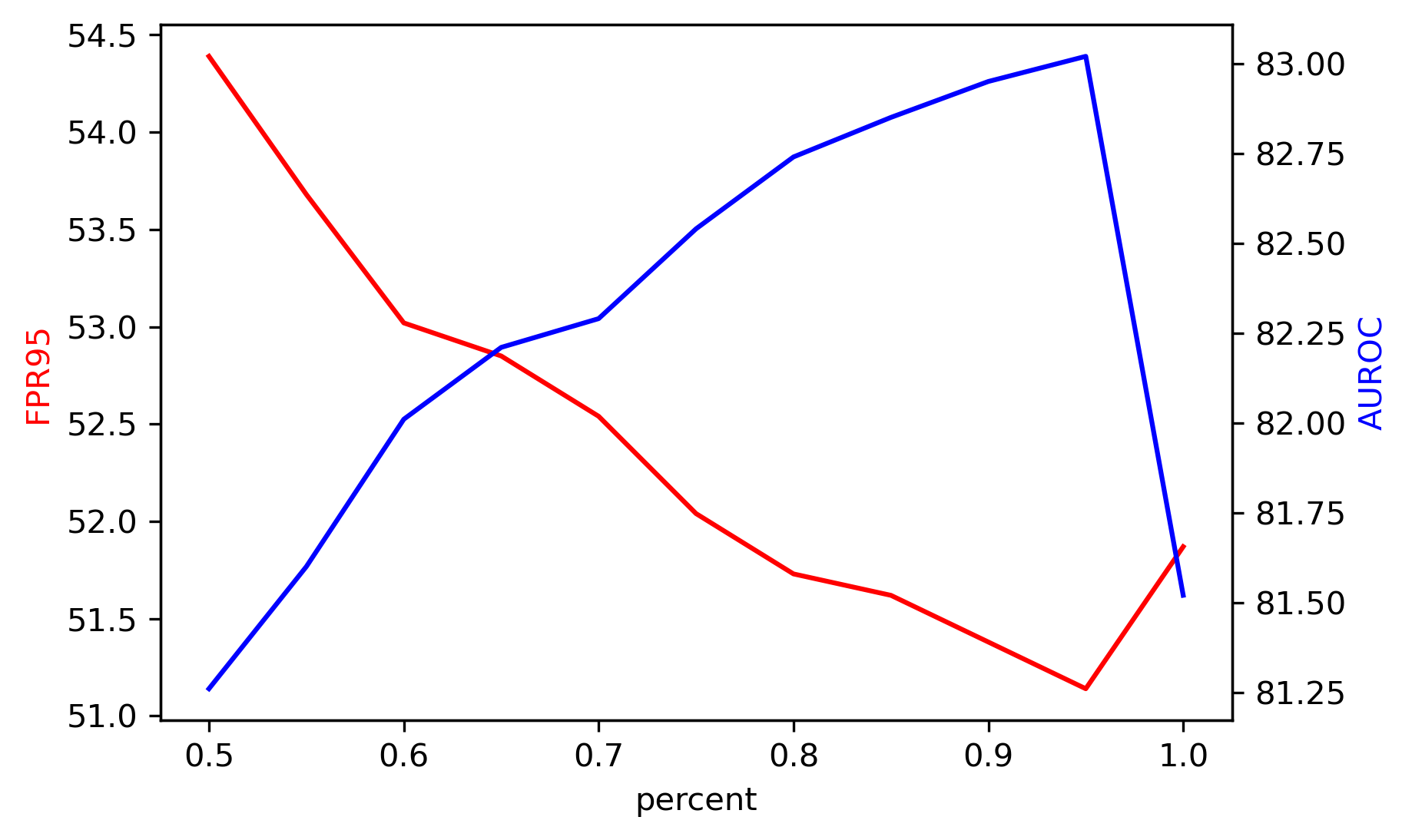}
\end{subfigure}
\vspace{-0.1in}
\caption{RLR results with different percentile of selected subset data on CIFAR-100 with MSP(left) and ODIN(right).}
\label{RLR:percentile}
\vspace{-0.15in}
\end{figure}

We plot the performance improvement of our RLR as a function of the percentile of chosen subset on CIFAR-100 over two base OOD scoring metrics MSP and ODIN, as shown in Fig ~\ref{RLR:percentile}. The percentile of chosen data varies from $0.5$ to $1$, with step $0.05$. When percentile is $1$, it is equivalent to our DLR. No matter with MSP or ODIN, when we decrease the percentile, the AUROC and FPR95 improve first, which means our RLR selects proper subsets for learning the regression, eliminating the noisy samples. When we further decrease the percentile, the performances start getting worse as the training data is too limited to learn a good regressor.

\subsubsection{Why RLR Failed on ImageNet-1k Benchmarks?}
\begin{figure}
\centering
\begin{subfigure}
\centering
\includegraphics[width=0.35\linewidth]{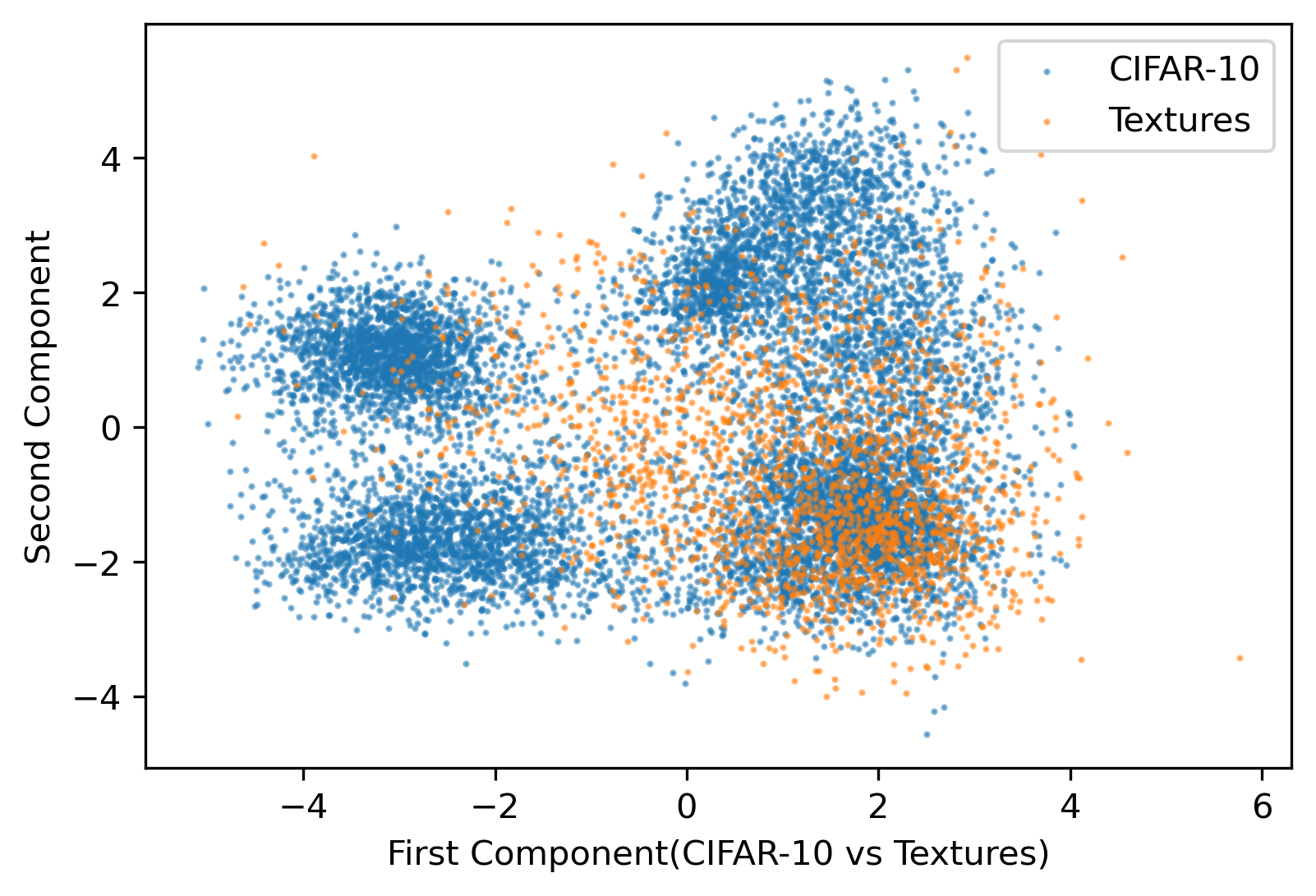}
\end{subfigure}
\begin{subfigure}
\centering
\includegraphics[width=0.35\linewidth]{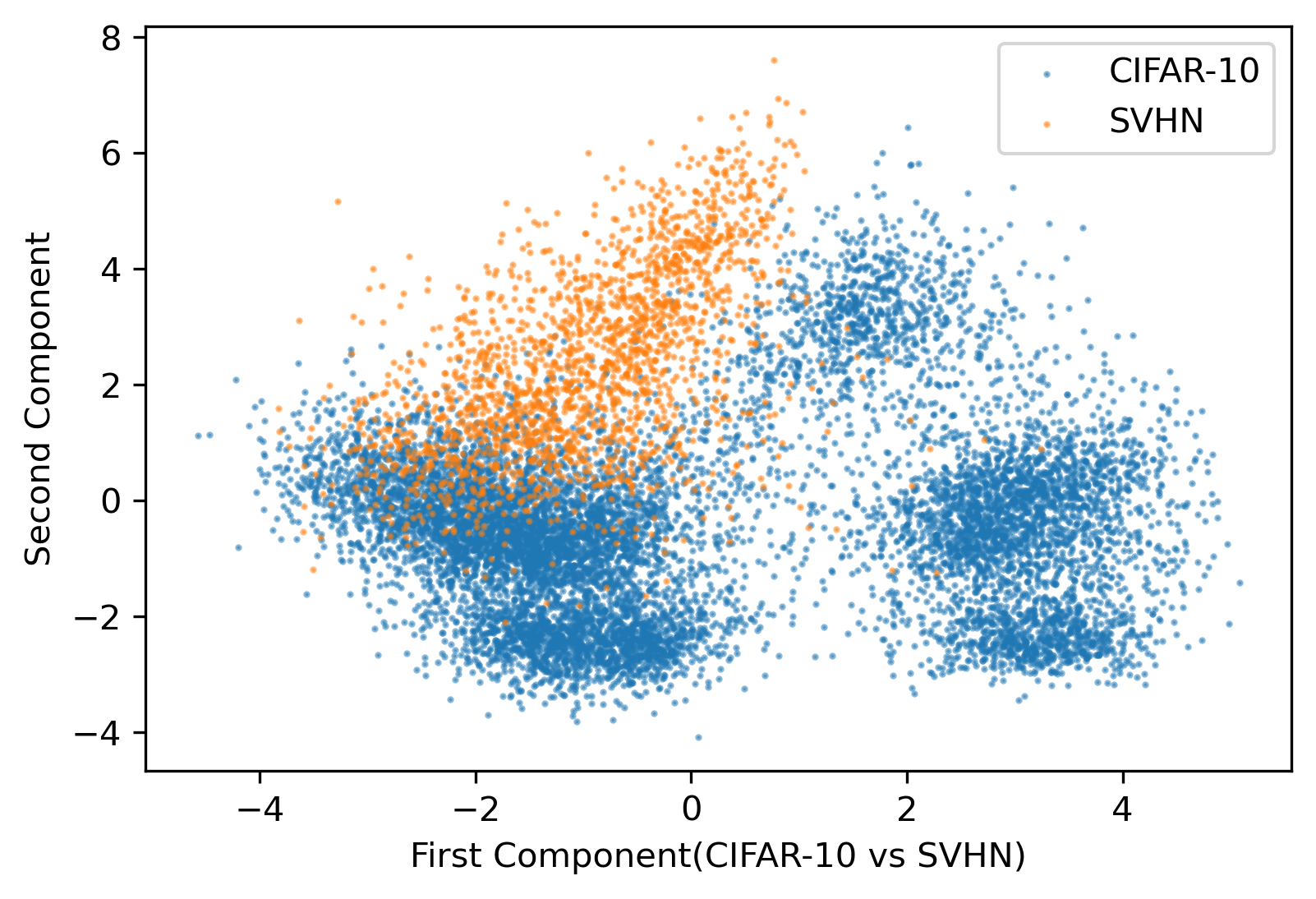}
\end{subfigure}
\begin{subfigure}
\centering
\includegraphics[width=0.35\linewidth]{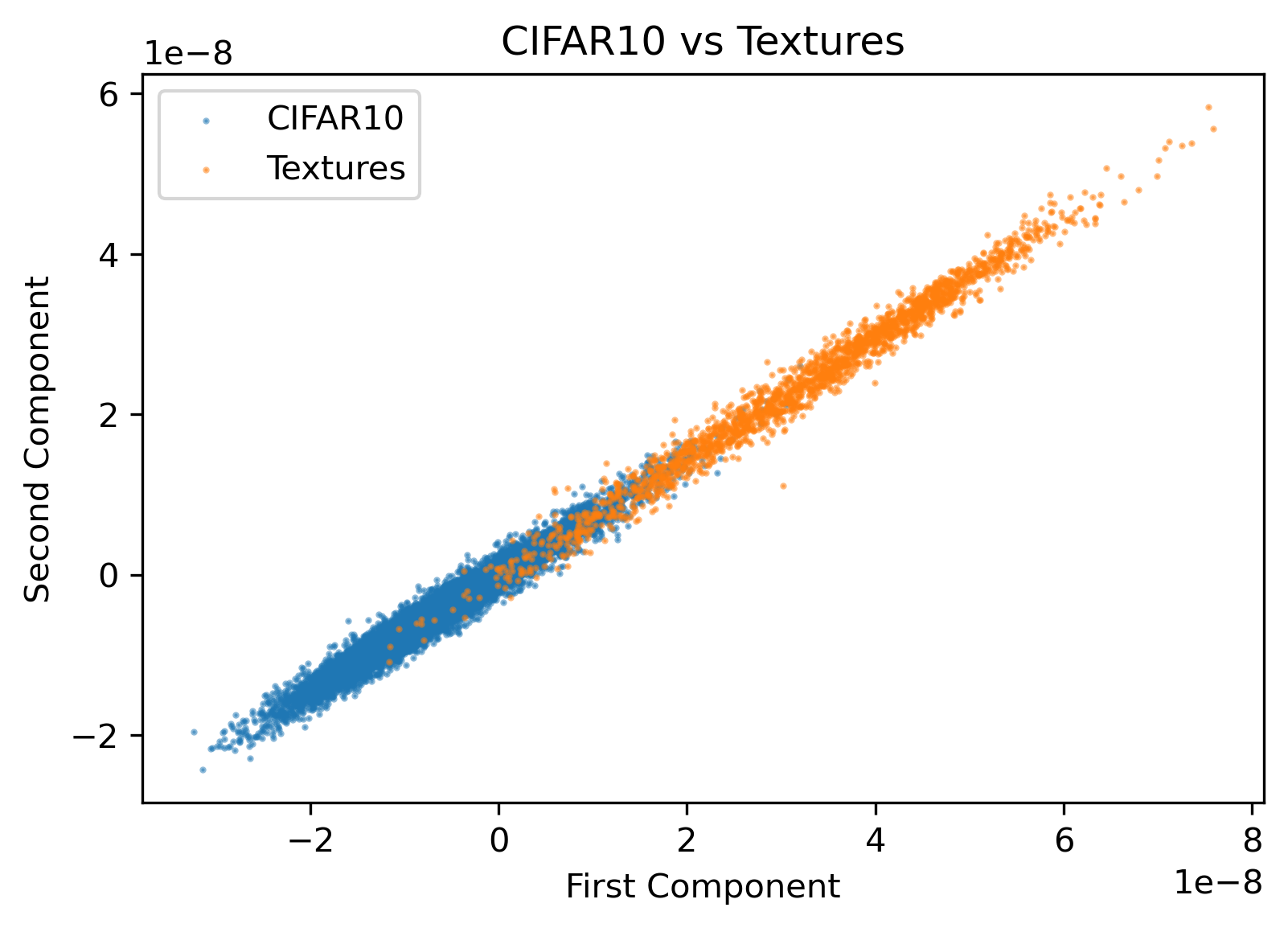}
\end{subfigure}
\begin{subfigure}
\centering
\includegraphics[width=0.35\linewidth]{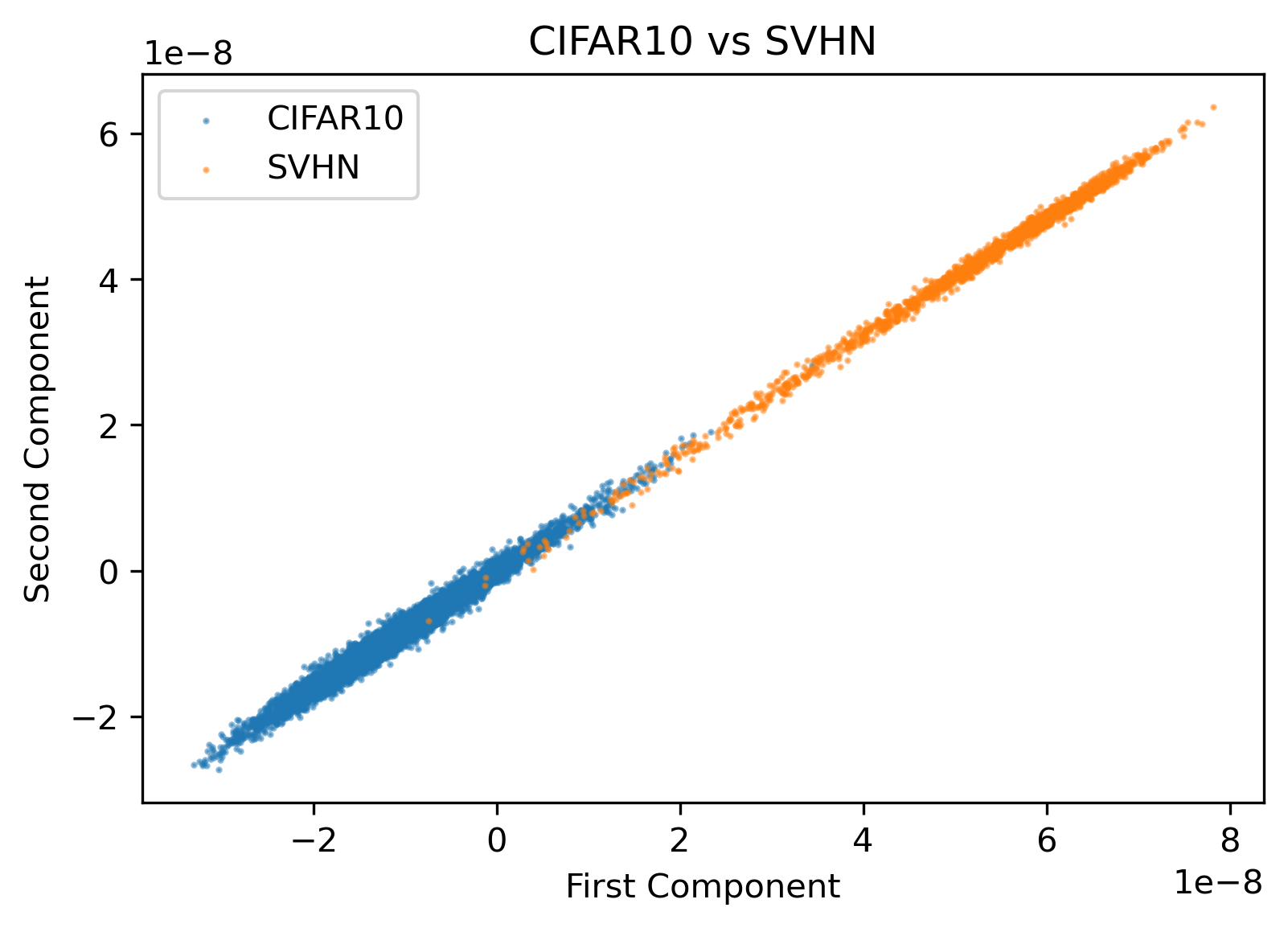}
\end{subfigure}
\caption{Visualization of feature distribution on CIFAR and OOD datasets.\label{CIFAR:feature}}
\vspace{-0.1in}
\end{figure}

Similar to Section~\ref{sec:linear separability}, we visualize the feature distribution on CIFAR and an OOD dataset with PCA in the first column of Fig.~\ref{CIFAR:feature}. We can see on CIFAR the separation of in- and out-of-distribution data is not as good as ImageNet-1k. This indicates that on ImageNet-1k benchmarks the feature and score linear relation is strong enough such that simple DLR can fit it well. On the contrary, filtering out subset data reduces the useful training data points, explaining why RLR failed to improve DLR in this case. However, we emphasize that the feature of CIFAR and OOD data is still roughly linearly separable, while PCA, as an unsupervised reduction method, may undermine linear separability. To better visualize the linear separability, we used LDA (with access to ground truth label) as an alternative method to reduce the dimension, and clearly display the linear separability in the second column of Fig.~\ref{CIFAR:feature}.

\subsubsection{Batch Size Effect of Online DLR}
We also check how the performance of our online DLR varies upon the batch size numbers. We run our online DLR with MSP in Table~\ref{tabel:ChangeOnlineBatchSize} by varying the batch size from $32$ to \emph{full} on CIFAR OOD benchmarks. From the results we can see that the performance of our online DLR is insensitive to the batch size number.

\begin{table}
\centering
\caption{Online DLR on CIFAR with different batch size.  \label{tabel:ChangeOnlineBatchSize}} 
\begin{tabular}{lcccc}
\toprule
In Dataset & \multicolumn{2}{c}{CIFAR-10} & \multicolumn{2}{c}{CIFAR-100}  \tabularnewline
Batch Size & FPR95$\downarrow$ & AUROC$\uparrow$ & FPR95$\downarrow$ & AUROC$\uparrow$ \tabularnewline
\midrule
\midrule
Raw Softmax & 51.37 & 90.87 & 80.21 & 75.67  \tabularnewline
32 & 15.06 & 96.05 & 55.84 & 83.31 \tabularnewline
64 & 15.04 & 96.07 & 55.77 & 83.34 \tabularnewline
128 & 14.99 & 96.10 & 55.64 & 83.38 \tabularnewline
256 & 14.79 & 96.24 & 55.29 & 83.53 \tabularnewline
512 & 14.53 & 96.38 & 54.79 & 83.69 \tabularnewline
1024 & 14.16 & 96.52 & 53.98 & 83.89 \tabularnewline
All data & \textbf{12.30} & \textbf{97.01} & \textbf{51.63} & \textbf{84.50}\tabularnewline
\bottomrule
\end{tabular}
\end{table}

\subsubsection{Combined with other methods} Recently, there are many method which can improve the OOD baseline detector. The most representative of these method is ReAct~\cite{sun2021react}. We find that our method is orthogonal to ReAct,  which means they can be combined. In Table.~\ref{noise_and_react} we can see our DLR outperforms ReAct with both MSP and ODIN. More interestingly, combining ReAct with DLR could get better result in some case.

\begin{table}
\caption{ImageNet-1k with ReAct.\label{noise_and_react}}
\begin{centering}
\begin{tabular}{lllcc}
\toprule
Method & ReAct & DLR & FPR$\downarrow$ & AUROC$\uparrow$ \tabularnewline
\midrule
\midrule
\multirow{4}{*}{MSP} & no & no & 76.98 & 79.28 \tabularnewline
& no & yes & \textbf{46.84} & 86.71 \tabularnewline
& yes & no & 70.19 & 81.73 \tabularnewline
& yes & yes & 67.83 & 81.98 \tabularnewline \midrule
\multirow{4}{*}{ODIN}& no & no & 72.99 & 82.56 \tabularnewline
& no & yes & 53.28 & 85.68 \tabularnewline
& yes & no & 63.64 & 84.49 \tabularnewline
& yes & yes & 48.85 & \textbf{87.97} \tabularnewline
\bottomrule
\end{tabular}
\par\end{centering}
\end{table}

\subsubsection{Revision Under different in-distribution rates}
\begin{figure}
\centering
\begin{subfigure}
\centering
\includegraphics[width=0.35\linewidth]{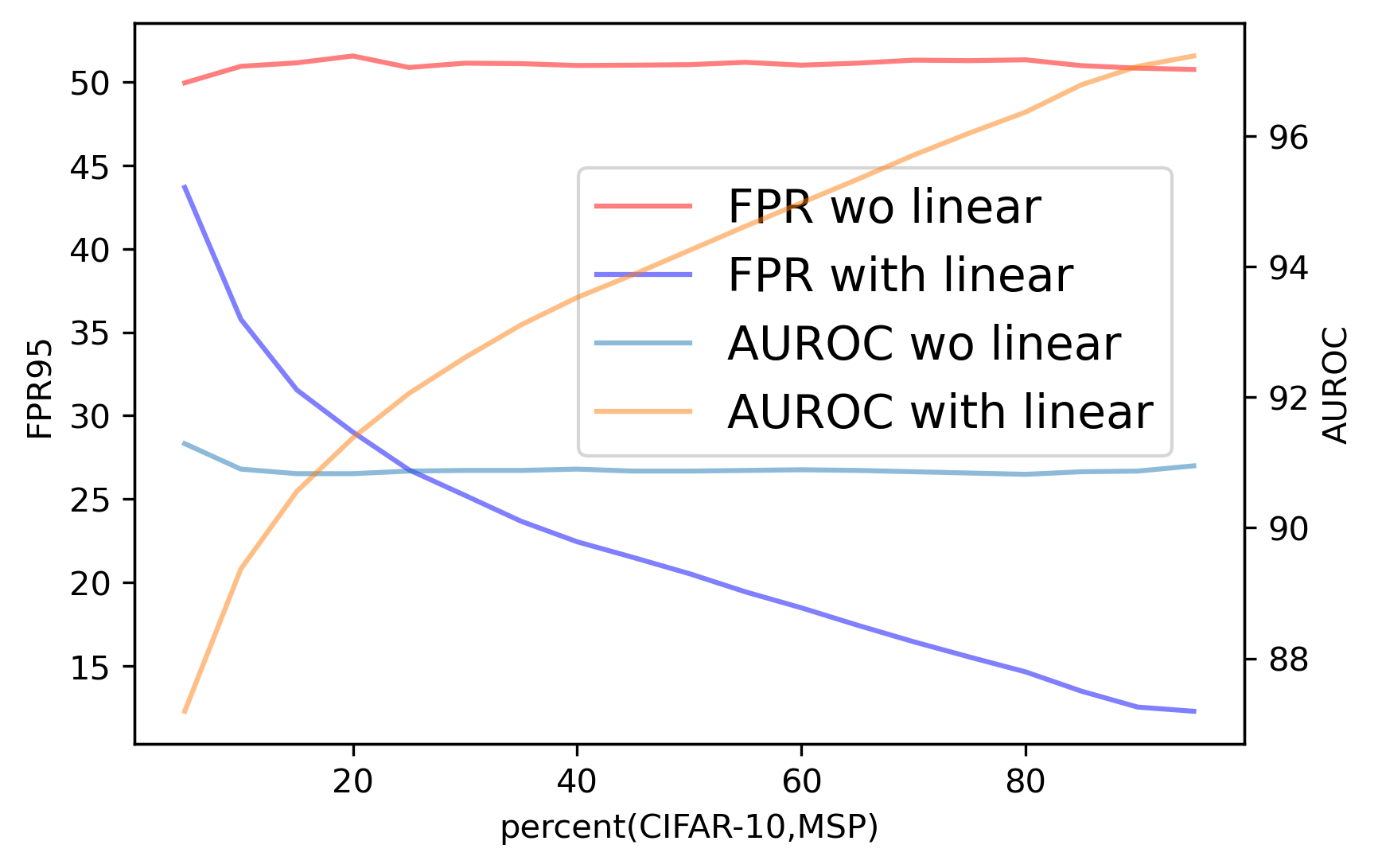}
\end{subfigure}
\begin{subfigure}
\centering
\includegraphics[width=0.35\linewidth]{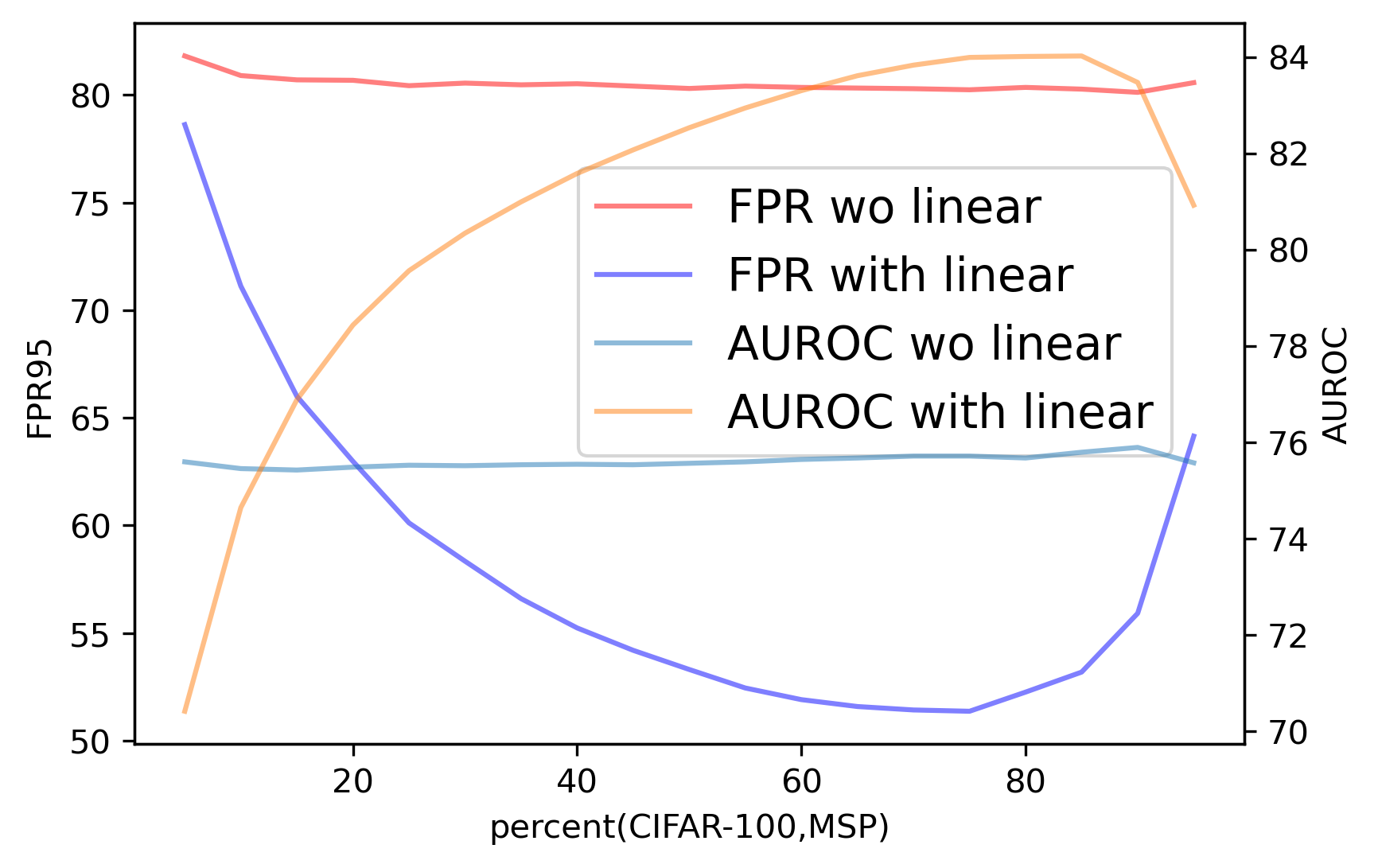}
\end{subfigure}
\vskip\baselineskip
\begin{subfigure}
\centering
\includegraphics[width=0.35\linewidth]{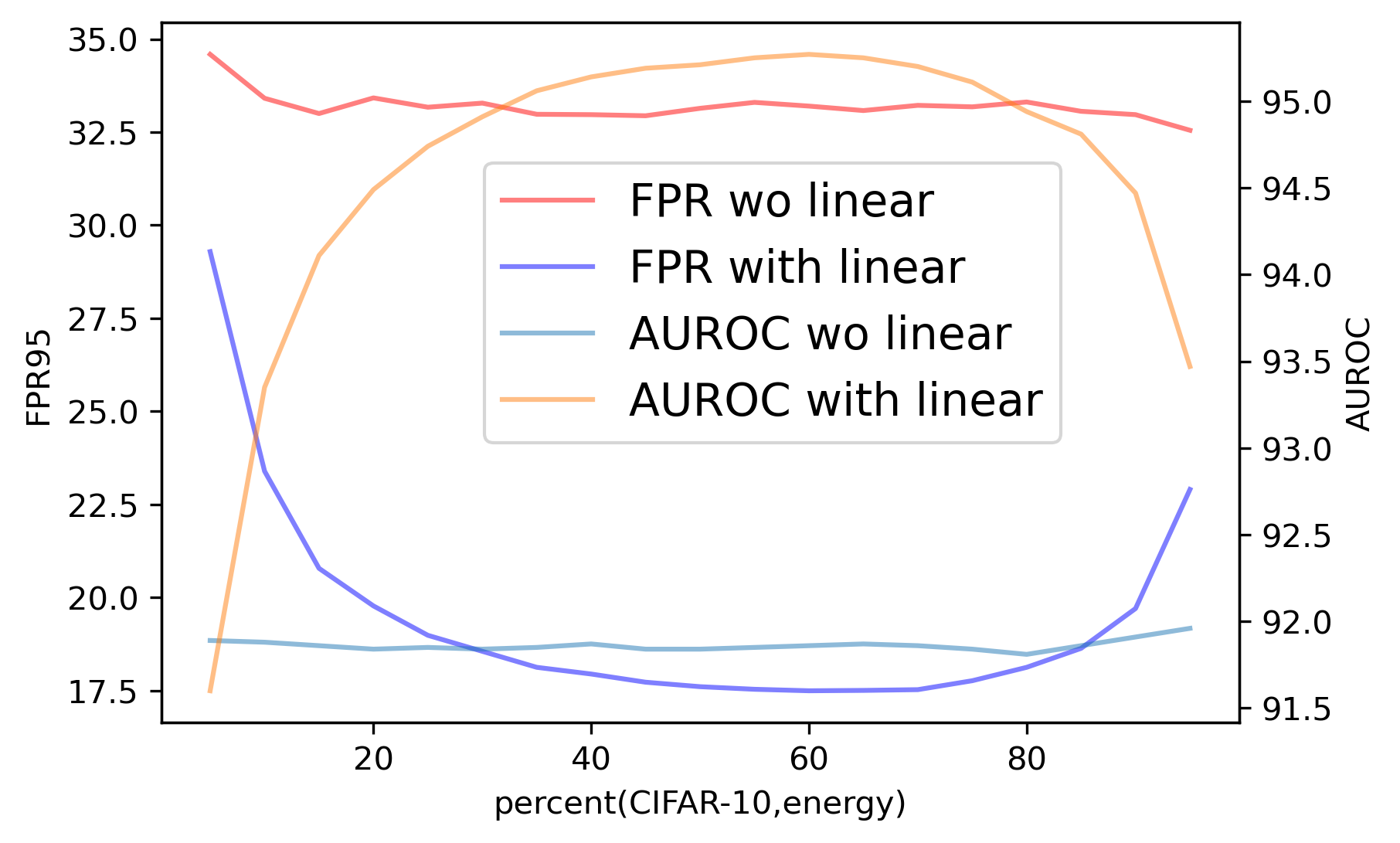}
\end{subfigure}
\begin{subfigure}
\centering
\includegraphics[width=0.35\linewidth]{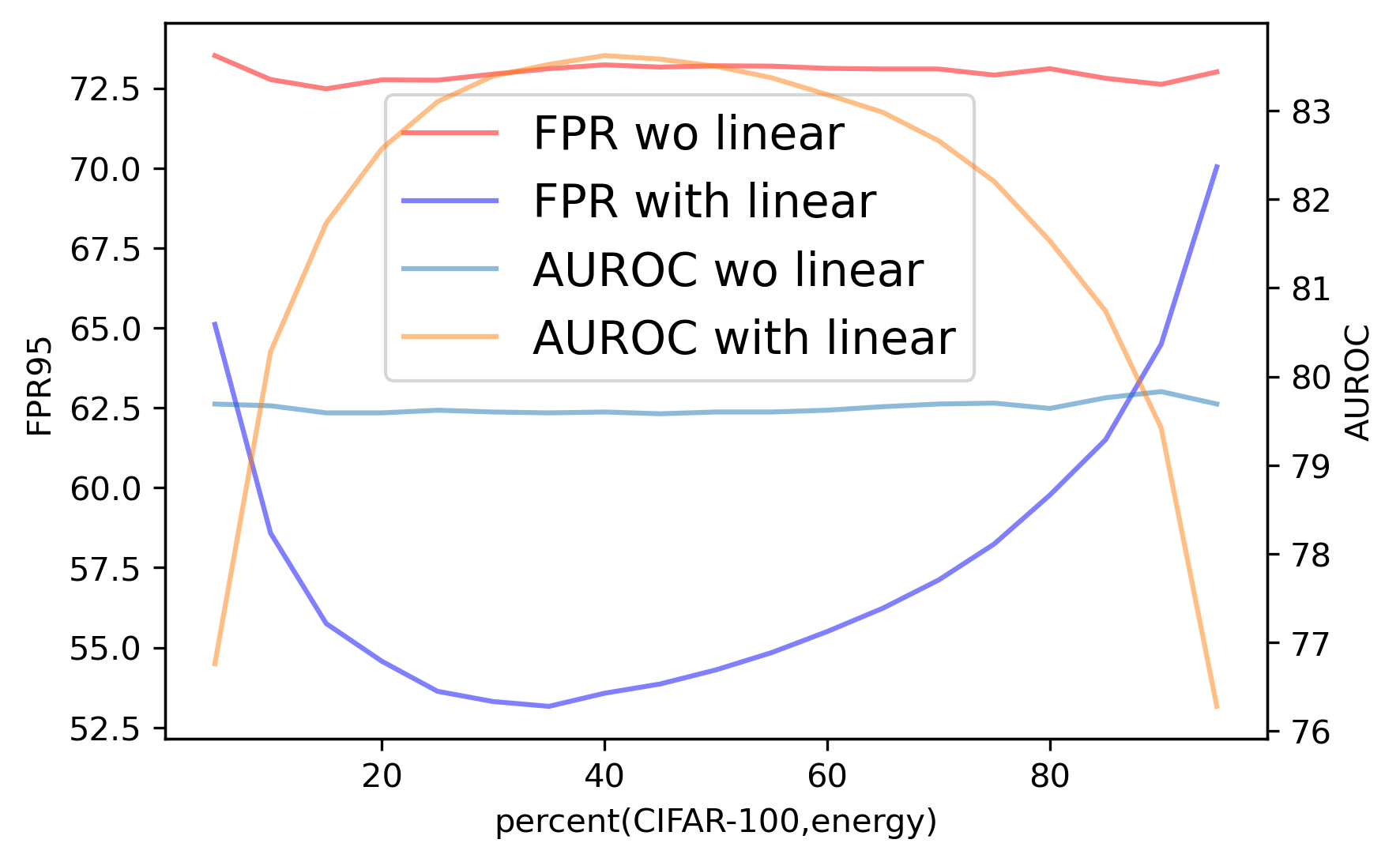}
\end{subfigure}
\vspace{-0.1in}
\caption{Results under different in-distribution rate\label{diff:in_rate}}
\end{figure}

From this section on, we only utilize DLR and focus on how linear calibration can improve the baseline OOD detector.
On CIFAR-10 and CIFAR-100, we tried to change the rate of in-distribution data for OOD detection. The rate of in-distribution data vary from $5\%$ to $95\%$ in Fig. ~\ref{diff:in_rate}, with spacing $5\%$. We fix the total number of data at 5000 because some out of distribution have less than 10000 images. Every experiment is repeated 10 times and the average is used as the final results.
We find that using MSP or energy as base OOD detector, our linear revision succeed in making correct classification with some improvements even with only $20\%$ to $40\%$ of in-distribution data. 
When the rate of in-distribution get smaller, our proposed linear revision will instead worsen the results. This is because when there is too small number of in-distribution, it is hard to capture the rich information of in-distribution. However, this is still astonishing because our methods still work when in-distribution only account for a small part of the all observation, where the classical outlier detection methos.  
We argue that this is because we utilized the information of pre-trained classifiers. Even if in-distribution data compose of the smaller part of dataset during OOD testing, OOD scores tend to produce larger scores for inliers and lower score for outliers, and linear regression will keep this tendency. 

\subsubsection{Revision Under different sample number}
\begin{figure}
\centering
\begin{subfigure}
\centering
\includegraphics[width=0.35\linewidth]{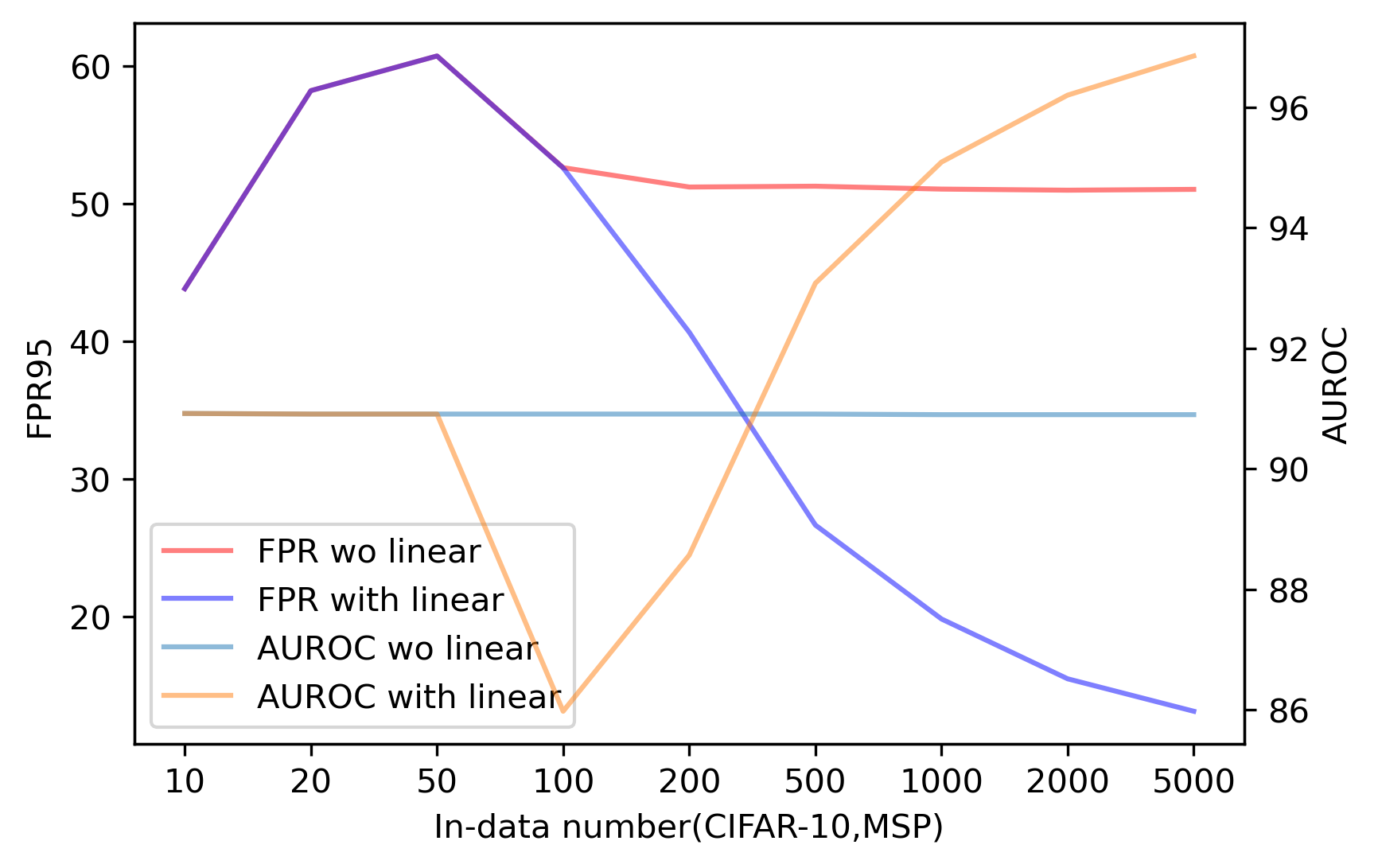}
\end{subfigure}
\begin{subfigure}
\centering
\includegraphics[width=0.35\linewidth]{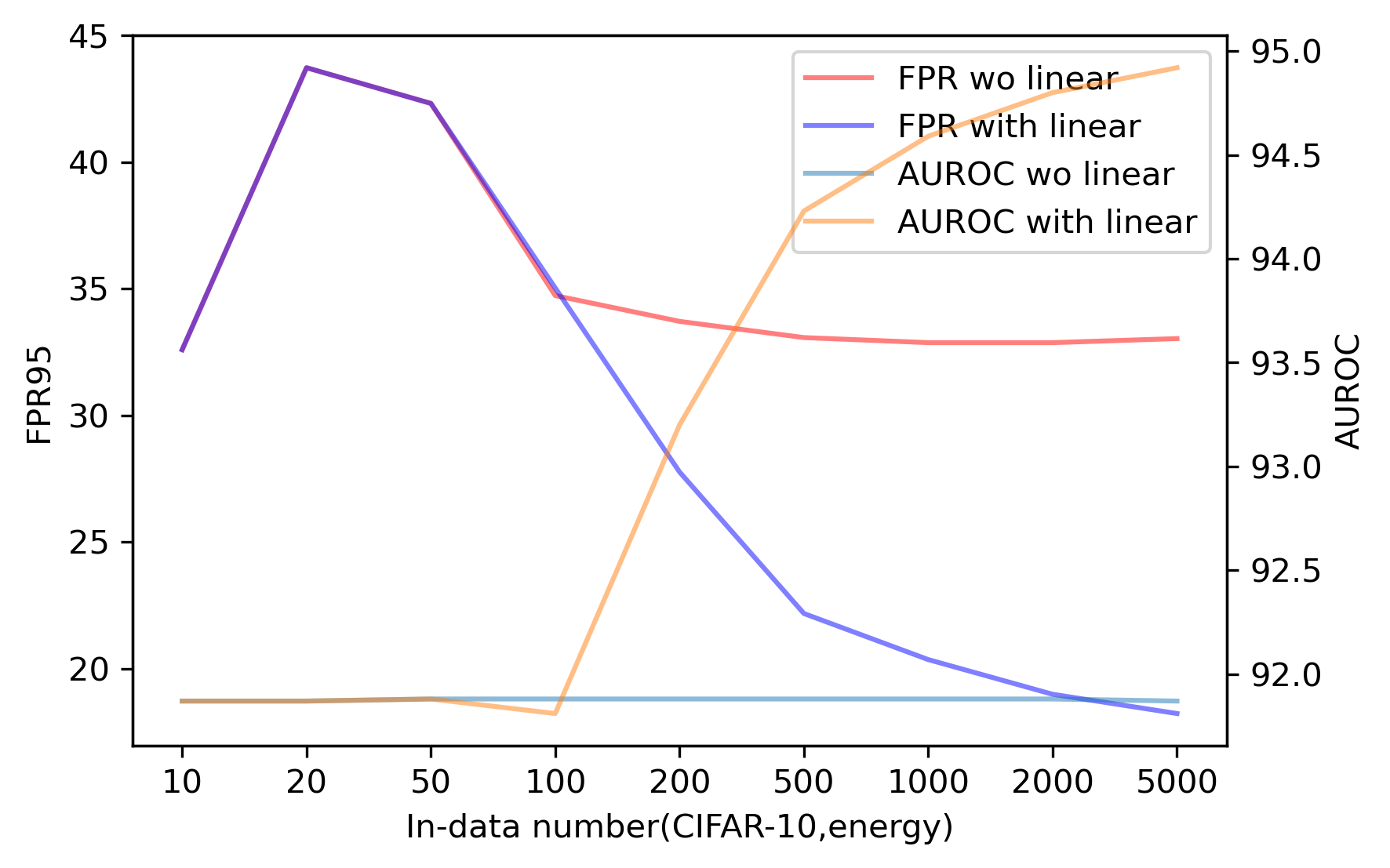}
\end{subfigure}
\vskip\baselineskip
\begin{subfigure}
\centering
\includegraphics[width=0.35\linewidth]{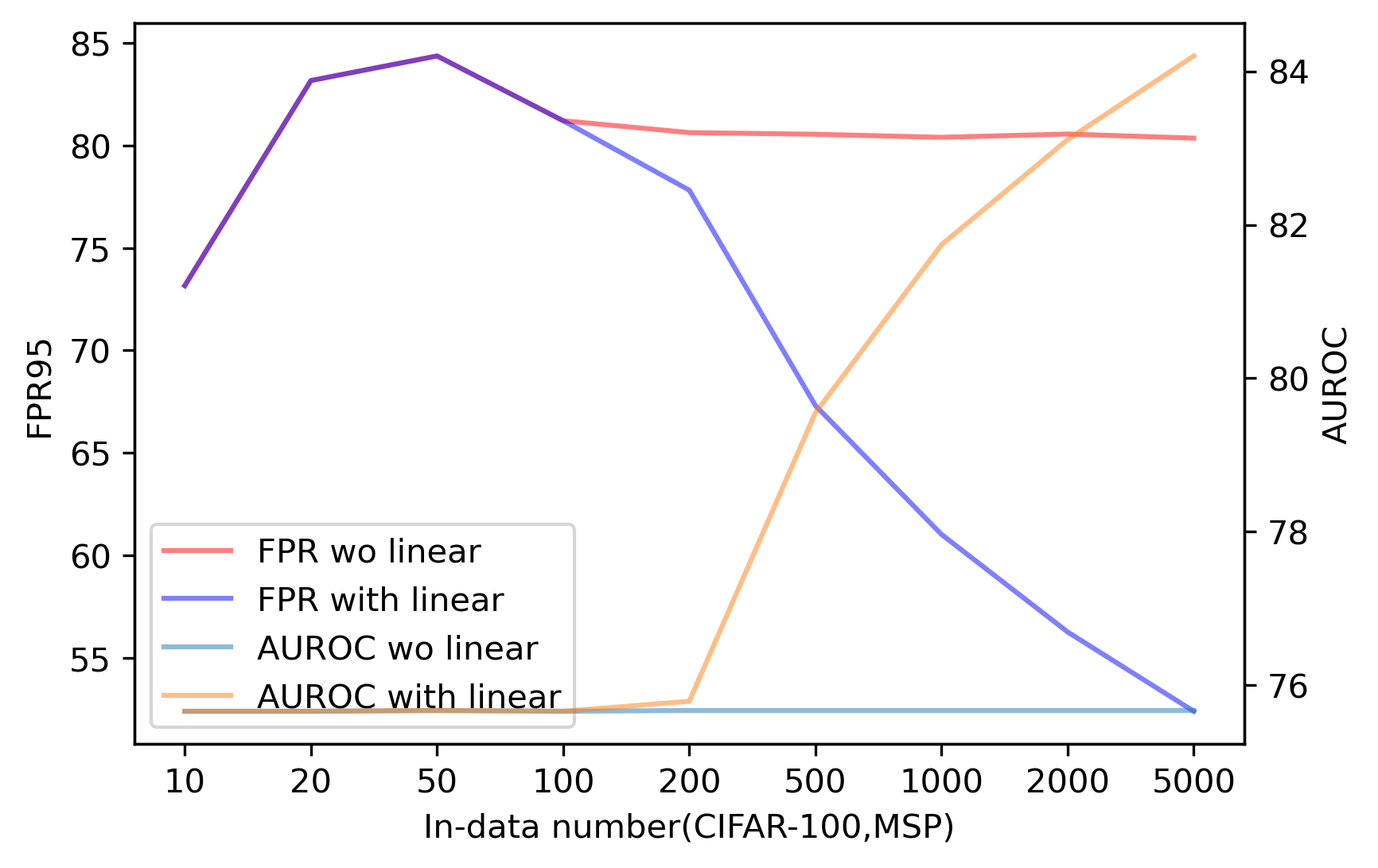}
\end{subfigure}
\begin{subfigure}
\centering
\includegraphics[width=0.35\linewidth]{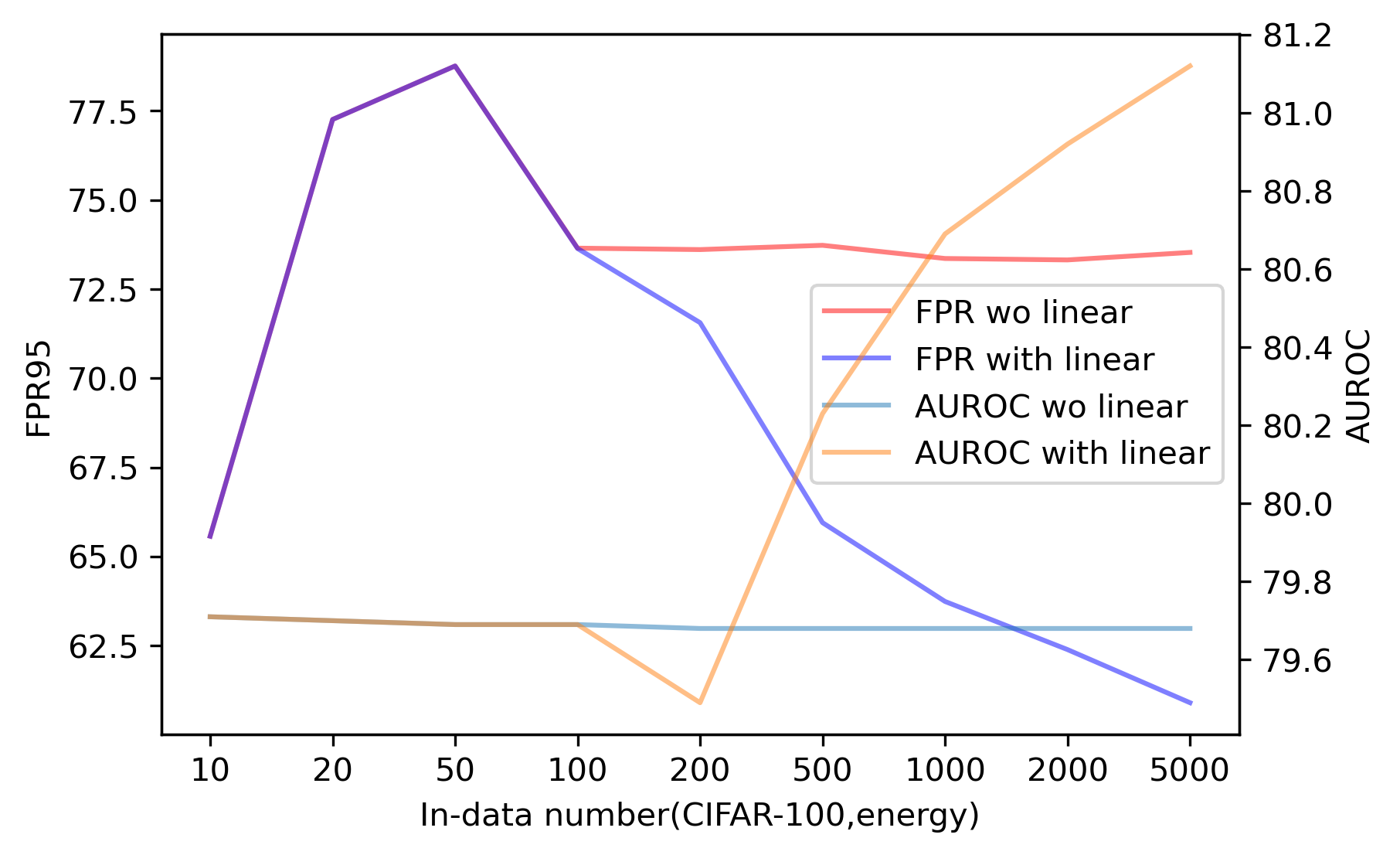}
\end{subfigure}
\vspace{-0.1in}
\caption{Results under different total sample number\label{diff_in_num}}
\end{figure}
Still on CIFAR, we keep the rate of in-distribution data and out-of-distribution data at 5:1, then change the total number of in-distribution from 10000 to 10 in Fig.~\ref{diff_in_num}. It is worth noting that we repeat more rounds when the total sample number is smaller to get more accurate estimation. Let's denote $m$ as the number of in-distribution data fed in the test time. We correspondingly repeat $\frac{10^5}{m}$ rounds. We find that when number of test data is too small (around the dimension of feature, note that the penultimate layer of Wide ResNet-40 of CIFAR use 128-dimensional), our algorithms will produce no revision because it fit the OOD score perfectly. When the number of data grow, our linear revision may first 
decease AUROC. However, our algorithm consistently improve the results when there are more than 500 in-distribution samples.

\subsubsection{Extend the results with multiple OOD data sets} 
\begin{figure}
\centering
\begin{subfigure}
\centering
\includegraphics[width=0.6\linewidth]{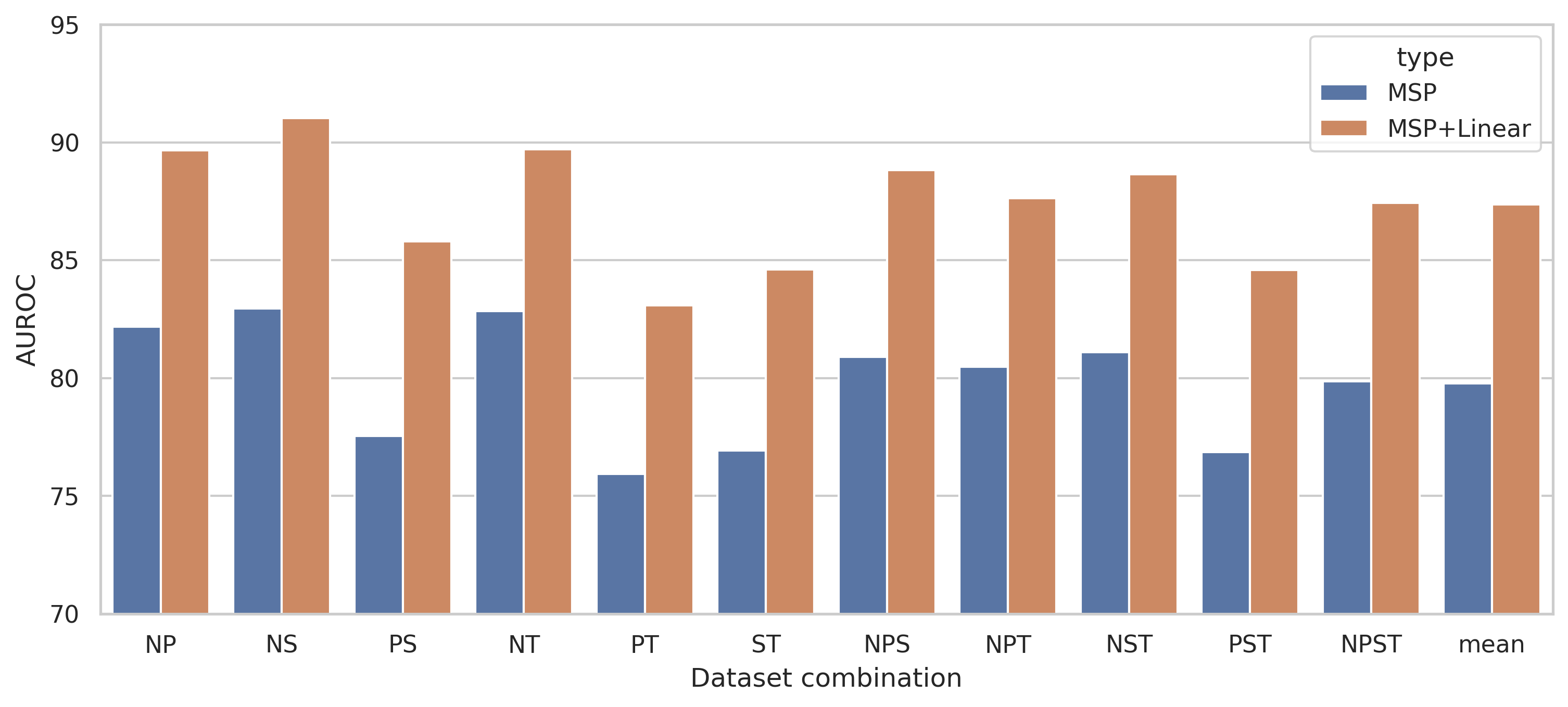}
\end{subfigure}
\begin{subfigure}
\centering
\includegraphics[width=0.6\linewidth]{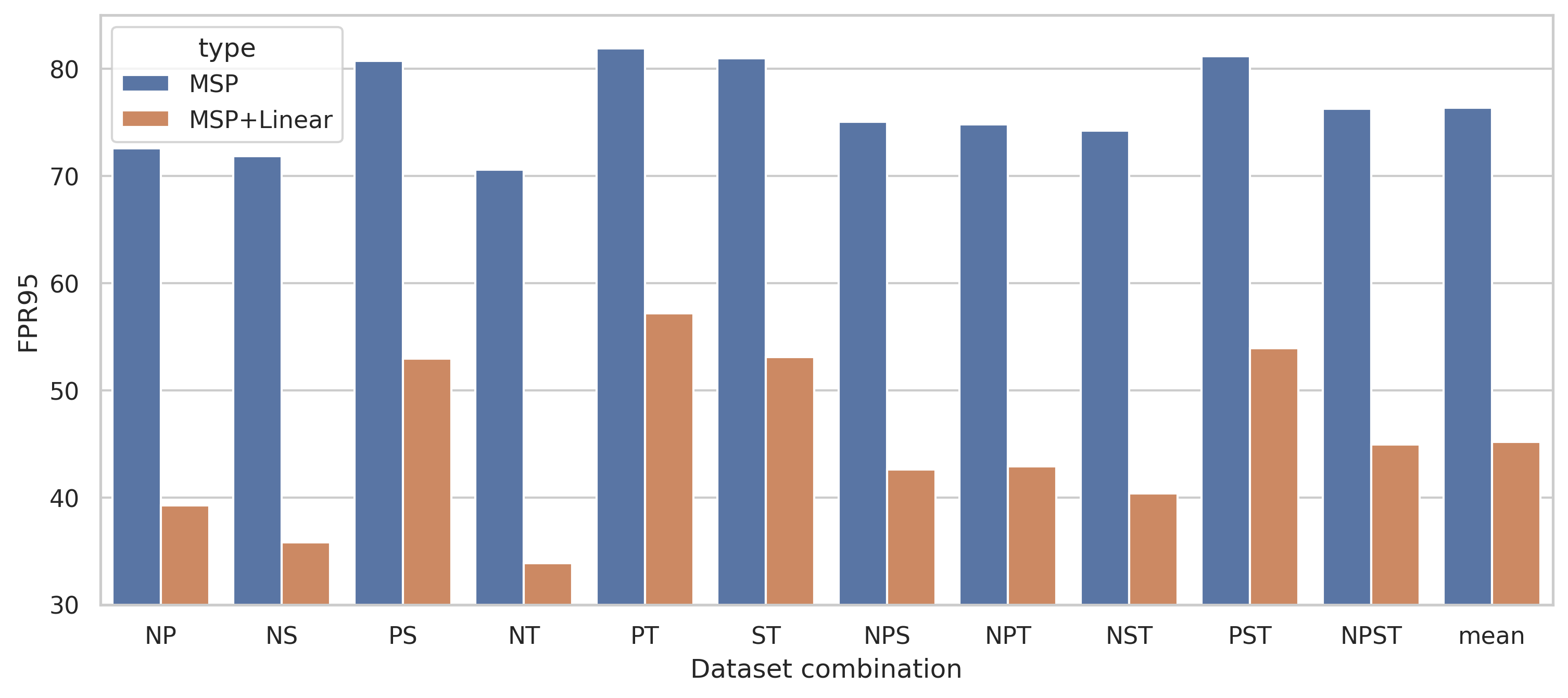}
\end{subfigure}
\vspace{-0.1in}
\caption{Results of Combination of Datasets\label{multiset}}
\end{figure}

Because all the above experiments carried out on single datasets then take average over all datasets, we wonder if our methods work when multi out-of-distribution domains is simultaneously given during. We carry out experiments on large scale OOD detection, fix the ImageNet-1k validation set as in-distribution. The out-of-distribution data are all the combination of i\textbf{N}aturalist, \textbf{P}laces, \textbf{S}UN and \textbf{T}exturs datasets (shorted as NPST). The results are visualized in Fig.~\ref{multiset} 
We find that our DLR can improve on all the combination of the given four datasets.
\subsubsection{Results on Noise}
Sometimes, noises are more challenging than disjointed datasets as OOD data. We also test how our DLR can help in these cases. We test two kind of noise, Gaussian noise with $\mu=0.5$ clipped to $\lbrack0, 1\rbrack$ and Uniform noise between $\lbrack0, 1\rbrack$. The results are shown in Tab.~\ref{noise}. Without DLR, both MSP and ODIN can not distinguish in- and out-of-distribution data easily. After applying DLR, the results greatly improve. 
\begin{table}
\caption{Results on Noise.\label{noise}}
\begin{centering}
\begin{tabular}{llcc}
\toprule
Method & Noise & FPR$\downarrow$ & AUROC$\uparrow$ \tabularnewline
\midrule
\midrule
\multirow{2}{*}{MSP} & Uniform & 99.60 & 89.13 \tabularnewline
& Gaussian & 99.77 & 88.08  \tabularnewline
\multirow{2}{*}{MSP+DLR}& Uniform & 0.01 & 99.50  \tabularnewline
& Gaussian & 0.01 & 99.34 \tabularnewline\midrule
\multirow{2}{*}{ODIN}& Uniform & 99.14 & 90.12  \tabularnewline
 & Gaussian & 99.56 & 89.15  \tabularnewline
\multirow{2}{*}{ODIN+DLR} & Uniform & 7.05 & 97.29\tabularnewline
& Gaussian & 15.38 & 96.73 \tabularnewline\midrule
\bottomrule
\end{tabular}
\par\end{centering}
\end{table}
\section{Conclusion}
In this paper, we find that a roughly linear relation between features extracted by deep neural networks and their OOD scores produced by those popular OOD detection algorithms on various standard OOD datasets.
Thus an embarrassingly simple test-time linear training model (ETLT) is proposed to discover and exploit this linear relation for OOD scores prediction directly based on the input feature, including DLR and a robust variant RLR.
Though simple, our ETLT methods outperform those popular OOD detection methods and achieve new the state-of-the-art performance on many benchmarks.

\clearpage
\bibliographystyle{unsrt}  
\bibliography{references} \clearpage
\begin{appendices}
\section{Experiments Setting}
For all results except for changing the percentile of chosen data in RLR, we set the regularization $\lambda = 1e-5$ and chosen percentile $p=80\%$. For experiments on CIFAR-10 and CIFAR-100, we directly use the feature of the penultimate layer of Wide ResNet-40. On ImageNet-1k, we further apply principal component analysis to the output of penultimate layer of ResNetv2-101, to reduce the dimension of feature to 64. For GMM model on CIFAR-10 and CIFAR-100, we use Gaussian mixtures of 10 and 100 components respectively, and each component has its own general covariance matrix. For GMM model on ImageNet-1k, a Gaussian Mixture of 1000 components with diagonal covariance matrix is used. The log-likelihood of each sample is used as an OOD score. For local outlier factor model, we set number of neighbors to 20. For Isolation Forest, the number of base estimators is set to 100.

\section{Detailed results on ImageNet-1k and CIFAR}
Here we list the detailed results of experiments on every single OOD datasets, which is omitted in the paper due to page limitation. The results of ImageNet-1k, CIFAR-10 and CIFAR-100 are displayed respectively in Tab.~\ref{Tab:imagenet},Tab.~\ref{Tab:cifar10},Tab.~\ref{Tab:cifar100}.
\begin{table*}[!ht]
\caption{Detailed Experiments on ImageNet-1k with AUPR\label{Tab:imagenet}}
\small
    \centering
    \begin{tabular}{l|ccc|ccc}
    \hline
        Dataset & \multicolumn{3}{c}{iNaturalist}  &\multicolumn{3}{c}{SUN} \tabularnewline
        metrics & FPR95 & AUROC & AUPR & FPR95 & AUROC & AUPR \tabularnewline\midrule\midrule
        MSP & 63.69 & 87.59 & 97.23 & 79.98 & 78.34 & 94.45 \tabularnewline
        +RLR & 21.84 & 94.87 & 98.69 & 53.27 & 86.46 & 96.44 \tabularnewline
        +DLR & 21.00 & 94.98 & 98.71 & 50.68 & 87.13 & 96.61\tabularnewline\midrule
        Energy & 64.91 & 88.48 & 97.58 & 65.33 & 85.32 & 96.57 \tabularnewline
        +RLR & 46.51 & 91.29 & 98.07 & 53.23 & 88.82 & 97.50 \tabularnewline
        +DLR & 45.48 & 91.04 & 97.96 & 52.11 & 88.67 & 97.38 \tabularnewline\midrule
        KL & 64.91 & 88.48 & 97.58 & 65.32 & 85.31 & 96.57 \tabularnewline
        +RLR & 46.50 & 91.29 & 98.07 & 53.23 & 88.82 & 97.50 \tabularnewline
        +DLR & 45.48 & 91.04 & 97.96 & 52.10 & 88.67 & 97.38 \tabularnewline\midrule
        ODIN & 62.69 & 89.36 & 97.76 & 71.67 & 83.92 & 96.26 \tabularnewline
        +RLR & 37.28 & 92.49 & 98.27 & 54.51 & 87.50 & 97.01 \tabularnewline
        +DLR & 34.84 & 92.94 & 98.37 & 51.31 & 92.94 & 97.26 \tabularnewline\midrule
        GradNorm & 50.03 & 90.33 & 97.83 & 46.48 & 89.03 & 97.29 \tabularnewline
        GMM & 87.90 & 68.43 & 90.26 & 89.99 & 63.29 & 88.72 \tabularnewline
        LOF & 95.16 & 51.57 & 83.87 & 94.89 & 52.27 & 84.04 \tabularnewline
        IF & 88.58 & 61.60 & 87.92 & 90.12 & 57.85 & 86.09 \tabularnewline\midrule\midrule
        Dataset & \multicolumn{3}{c}{Places}  &\multicolumn{3}{c}{Textures} \tabularnewline
        metrics & FPR95 & AUROC & AUPR & FPR95 & AUROC & AUPR \tabularnewline\midrule\midrule
        MSP & 81.44 & 76.76 & 94.15 & 82.73 & 74.45 & 95.65 \tabularnewline
        +RLR & 59.06 & 83.79 & 95.75 & 59.02 & 80.01 & 96.22 \tabularnewline
        +DLR & 57.16 & 84.47 & 95.94 & 58.48 & 80.24 & 96.30 \tabularnewline\midrule
        Energy & 73.02 & 81.37 & 95.49 & 80.87 & 75.79 & 96.05 \tabularnewline
        +RLR & 64.42 & 83.73 & 96.01 & 76.49 & 73.76 & 95.57 \tabularnewline
        +DLR & 62.71 & 84.35 & 96.14 & 69.49 & 75.39 & 95.47 \tabularnewline\midrule
        KL & 73.02 & 81.37 & 95.49 & 80.87 & 75.79 & 96.05 \tabularnewline
        +RLR & 64.42 & 83.73 & 96.01 & 76.49 & 73.76 & 95.57 \tabularnewline
        +DLR & 62.71 & 84.35 & 96.14 & 69.49 & 75.40 & 95.47 \tabularnewline\midrule
        ODIN & 76.27 & 80.67 & 95.35 & 81.31 & 76.30 & 96.12 \tabularnewline
        +RLR & 62.87 & 83.48 & 95.84 & 66.95 & 76.36 & 95.58 \tabularnewline
        +DLR & 60.54 & 84.47 & 96.11 & 66.44 & 76.85 & 95.71 \tabularnewline\midrule
        GradNorm & 60.86 & 84.82 & 96.26 & 61.42 & 81.07 & 96.96 \tabularnewline
        GMM & 96.85 & 52.83 & 84.54 & 95.37 & 35.34 & 83.83 \tabularnewline
        LOF & 93.05 & 56.37 & 85.64 & 82.02 & 65.39 & 92.77 \tabularnewline
        IF & 93.45 & 50.24 & 83.02 & 54.34 & 87.76 & 98.27 \tabularnewline
        \bottomrule
    \end{tabular}
\end{table*}

\begin{table*}[!ht]
\caption{Details of Experiments On CIFAR-10\label{Tab:cifar10}}
    \centering
    \begin{tabular}{l|ccc|ccc|ccc}
    \hline
        Dataset &\multicolumn{3}{c}{Textures}&\multicolumn{3}{c}{SVHN} & \multicolumn{3}{c}{Places365}\tabularnewline
        Method & FPR & AUROC & AUPR & FPR & AUROC & AUPR & FPR & AUROC & AUPR \tabularnewline\midrule\midrule
        MSP & 59.50 & 88.37 & 97.16 & 48.98 & 91.86 & 98.26 & 60.32 & 88.08 & 97.08\tabularnewline
        +RLR & 27.35 & 91.89 & 97.33 & 11.24 & 95.91 & 98.46 & 31.49 & 93.22 & 98.32\tabularnewline
        +DLR & 20.83 & 94.56 & 98.43 & 10.47 & 96.71 & 98.96 & 29.86 & 93.46 & 98.40\tabularnewline\midrule
        Energy & 52.33 & 85.36 & 95.48 & 35.49 & 91.15 & 97.72 & 40.16 & 89.75 & 97.25\tabularnewline
        +RLR & 34.74 & 88.52 & 95.97 & 10.28 & 97.38 & 99.32 & 35.82 & 91.33 & 97.76\tabularnewline
        +DLR & 36.75 & 87.23 & 95.44 & 14.64 & 95.73 & 98.80 & 37.55 & 90.76 & 97.64\tabularnewline\midrule
        Odin & 49.62 & 84.57 & 95.14 & 32.88 & 92.11 & 98.03 & 57.14 & 84.23 & 95.73 \tabularnewline
        +RLR & 53.33 & 77.01 & 91.57 & 27.00 & 88.96 & 95.87 & 55.31 & 84.24 & 95.71\tabularnewline
        +DLR & 60.74 & 72.08 & 89.53 & 27.18 & 89.31 & 96.10 & 63.96 & 80.08 & 94.40\tabularnewline\midrule
        KL & 52.34 & 85.36 & 95.48 & 35.49 & 91.15 & 97.72 & 40.17 & 89.75 & 97.25\tabularnewline
        +RLR & 34.69 & 88.53 & 95.98 & 10.29 & 97.39 & 99.32 & 35.74 & 91.33 & 97.76\tabularnewline
        +DLR & 36.74 & 87.23 & 95.44 & 14.67 & 95.74 & 98.80 & 37.55 & 90.75 & 97.63\tabularnewline\midrule
        GradNorm & 73.59 & 57.90 & 83.12 & 59.49 & 70.21 & 89.45 & 78.38 & 60.51 & 86.99\tabularnewline
        GMM & 60.24 & 83.07 & 95.80 & 98.01 & 23.62 & 70.12 & 80.85 & 79.14 & 94.97\tabularnewline
        IF & 99.35 & 21.64 & 70.19 & 78.71 & 67.98 & 89.67 & 91.94 & 55.09 & 85.27\tabularnewline
        LOF & 96.90 & 45.94 & 80.20 & 92.86 & 57.39 & 86.12 & 98.62 & 52.92 & 84.58\tabularnewline\midrule\midrule
                Dataset & \multicolumn{3}{c}{LSUN-C} & \multicolumn{3}{c}{LSUN-R} &\multicolumn{3}{c}{iSUN}\tabularnewline
        Method & FPR & AUROC & AUPR & FPR & AUROC & AUPR & FPR & AUROC & AUPR \tabularnewline\midrule
        MSP & 30.95 & 95.63 & 99.13 & 52.23 & 91.49 & 98.17 & 56.24 & 89.80 & 97.73 \tabularnewline
        +RLR & 0.50 & 99.83 & 99.96 & 2.08 & 99.51 & 99.89 & 8.33 & 98.17 & 99.57 \tabularnewline
        +DLR & 0.18 & 99.92 & 99.98 & 3.73 & 99.19 & 99.82 & 8.75 & 98.19 & 99.59 \tabularnewline\midrule
        Energy & 8.31 & 98.34 & 99.65 & 27.75 & 94.15 & 98.65 & 33.84 & 92.51 & 98.23 \tabularnewline
        +RLR & 0.56 & 99.81 & 99.94 & 5.79 & 98.81 & 99.74 & 9.69 & 97.98 & 99.55 \tabularnewline
        +DLR & 1.44 & 99.61 & 99.89 & 7.05 & 98.58 & 99.69 & 10.62 & 97.87 & 99.54 \tabularnewline\midrule
        Odin & 15.90 & 96.98 & 99.33 & 26.63 & 94.58 & 98.77 & 32.45 & 93.29 & 98.48 \tabularnewline
        +RLR & 39.79 & 84.25 & 94.56 & 17.82 & 94.81 & 98.53 & 21.46 & 94.14 & 98.40 \tabularnewline
        +DLR & 37.13 & 85.77 & 95.23 & 19.65 & 94.32 & 98.41 & 18.96 & 95.08 & 98.69 \tabularnewline\midrule
        KL & 8.31 & 98.34 & 99.65 & 27.75 & 94.15 & 98.65 & 33.84 & 92.51 & 98.23 \tabularnewline
        +RLR & 0.56 & 99.81 & 99.94 & 5.81 & 98.81 & 99.74 & 9.67 & 97.98 & 99.55 \tabularnewline
        +DLR & 1.42 & 99.61 & 99.89 & 7.02 & 98.58 & 99.69 & 10.62 & 97.87 & 99.54 \tabularnewline\midrule
        GradNorm & 12.07 & 96.85 & 99.18 & 65.27 & 73.38 & 92.06 & 70.27 & 71.07 & 91.41\tabularnewline
        GMM & 96.12 & 46.31 & 80.17 & 95.84 & 58.68 & 89.29 & 95.10 & 58.88 & 89.14 \tabularnewline
        IF & 59.47 & 82.96 & 95.11 & 71.85 & 76.06 & 93.43 & 78.44 & 71.09 & 91.75\tabularnewline
        LOF & 97.62 & 51.69 & 83.89 & 94.21 & 65.72 & 90.49 & 94.64 & 65.04 & 90.08\tabularnewline
        \bottomrule
    \end{tabular}
\end{table*}

\begin{table*}[!ht]
\caption{Details of Experiments On CIFAR-100\label{Tab:cifar100}}
    \centering
    \begin{tabular}{l|ccc|ccc|ccc}
    \hline
        Dataset & \multicolumn{3}{c}{Textures} & \multicolumn{3}{c}{SVHN}  &\multicolumn{3}{c}{Places365}\tabularnewline
        Method & FPR & AUROC & AUPR & FPR & AUROC & AUPR & FPR & AUROC & AUPR \tabularnewline\midrule\midrule
        MSP & 83.55 & 73.67 & 93.07 & 84.00 & 71.44 & 92.89 & 82.30 & 74.03 & 93.30\tabularnewline
        +RLR & 57.25 & 80.62 & 93.02 & 44.29 & 84.18 & 94.93 & 76.63 & 76.16 & 93.53\tabularnewline
        +DLR & 64.26 & 78.55 & 92.87 & 56.75 & 78.72 & 93.42 & 75.19 & 76.47 & 93.59 \tabularnewline\midrule
        Energy & 79.32 & 76.40 & 93.70 & 85.43 & 73.96 & 93.62 & 80.11 & 75.80 & 93.59 \tabularnewline
        +RLR & 69.14 & 74.68 & 91.39 & 63.51 & 77.91 & 93.67 & 77.56 & 75.87 & 93.51 \tabularnewline
        +DLR & 70.08 & 73.51 & 91.29 & 64.85 & 78.73 & 94.23 & 80.46 & 73.55 & 92.84 \tabularnewline\midrule
        Odin & 79.57 & 73.43 & 92.81 & 87.73 & 65.43 & 90.95 & 87.14 & 72.00 & 92.67 \tabularnewline
        +RLR & 69.01 & 76.09 & 92.60 & 77.75 & 68.83 & 90.10 & 86.30 & 70.30 & 91.79 \tabularnewline
        +DLR & 65.44 & 74.61 & 92.13 & 67.50 & 69.81 & 89.50 & 88.29 & 68.33 & 91.21 \tabularnewline\midrule
        KL & 79.32 & 76.40 & 93.70 & 85.43 & 73.96 & 93.62 & 80.11 & 75.80 & 93.59 \tabularnewline
        +RLR & 69.14 & 74.68 & 91.39 & 63.51 & 77.91 & 93.67 & 77.56 & 75.87 & 93.51\tabularnewline
        +DLR & 70.08 & 73.51 & 91.29 & 64.85 & 78.73 & 94.23 & 80.46 & 73.55 & 92.84 \tabularnewline\midrule
        GradNorm & 87.48 & 60.41 & 87.46 & 97.30 & 55.00 & 86.83 & 96.95 & 53.65 & 85.83 \tabularnewline
        GMM & 92.18 & 63.26 & 89.34 & 99.09 & 62.07 & 91.35 & 86.33 & 73.32 & 92.46 \tabularnewline
        IF & 95.44 & 49.11 & 82.42 & 79.60 & 72.67 & 92.35 & 87.76 & 62.94 & 88.68 \tabularnewline
        LOF & 98.47 & 42.35 & 80.32 & 98.08 & 44.98 & 83.38 & 99.26 & 38.93 & 80.28\tabularnewline\midrule\midrule
                Dataset & \multicolumn{3}{c}{LSUN-C} &\multicolumn{3}{c}{LSUN-R} &\multicolumn{3}{c}{iSUN}\tabularnewline
        Method & FPR & AUROC & AUPR & FPR & AUROC & AUPR & FPR & AUROC & AUPR \tabularnewline\midrule\midrule
        MSP & 66.00 & 83.85 & 96.35 & 82.48 & 75.39 & 94.08 & 82.93 & 75.62 & 94.10\tabularnewline
        +RLR & 20.12 & 96.09 & 99.14 & 33.99 & 92.43 & 98.21 & 30.10 & 92.88 & 98.22\tabularnewline
        +DLR & 19.68 & 96.18 & 99.15 & 48.24 & 88.13 & 97.11 & 45.66 & 88.94 & 97.28\tabularnewline\midrule
        Energy & 35.78 & 93.46 & 98.60 & 79.14 & 79.42 & 95.04 & 81.00 & 78.99 & 94.92\tabularnewline
        +RLR & 25.53 & 94.94 & 98.86 & 56.23 & 85.66 & 96.46 & 56.42 & 85.53 & 96.40\tabularnewline
        +DLR & 27.81 & 93.83 & 98.51 & 59.68 & 83.83 & 95.98 & 60.90 & 83.43 & 95.85\tabularnewline\midrule
        Odin & 57.38 & 87.05 & 97.07 & 69.27 & 82.48 & 95.81 & 66.22 & 83.01 & 95.85\tabularnewline
        +RLR & 29.64 & 94.12 & 98.67 & 23.89 & 93.72 & 98.32 & 23.78 & 93.36 & 98.10\tabularnewline
        +DLR & 34.32 & 92.95 & 98.37 & 28.50 & 91.66 & 97.67 & 27.18 & 91.79 & 97.57\tabularnewline\midrule
        KL & 35.78 & 93.46 & 98.60 & 79.14 & 79.42 & 95.04 & 81.00 & 78.99 & 94.92\tabularnewline
        +RLR & 25.53 & 94.94 & 98.86 & 56.23 & 85.66 & 96.46 & 56.42 & 85.53 & 96.40\tabularnewline
        +DLR & 27.80 & 93.83 & 98.51 & 59.68 & 83.83 & 95.98 & 60.90 & 83.43 & 95.85\tabularnewline\midrule
        GradNorm & 39.47 & 92.12 & 98.22 & 99.06 & 40.06 & 80.11 & 99.06 & 44.09 & 82.10\tabularnewline
        GMM & 92.69 & 68.75 & 91.34 & 96.17 & 78.01 & 95.34 & 97.93 & 74.34 & 94.19\tabularnewline
        IF & 40.69 & 90.68 & 97.72 & 90.66 & 60.69 & 88.15 & 91.32 & 60.79 & 88.44\tabularnewline
        LOF & 97.37 & 36.64 & 77.84 & 98.02 & 49.36 & 85.27 & 98.16 & 47.67 & 84.65\tabularnewline
        \bottomrule
    \end{tabular}
\end{table*}
\end{appendices}

\end{document}